%% file: main.tex
\documentclass[runningheads]{llncs}

\usepackage{eccv}

\usepackage{placeins}
\usepackage{eccvabbrv}
\usepackage{multirow}
\usepackage{graphicx}
\usepackage{booktabs}
\usepackage{tablefootnote}

\usepackage{tikz}
\usetikzlibrary{arrows.meta}
\usetikzlibrary{matrix}
\usetikzlibrary{calc}

\newcommand{\commentsection}[1]{}

\usepackage[accsupp]{axessibility}  

\newcommand{\cropleft}[2][]{%
    \includegraphics[trim=180px 0px 40px 0px, clip, #1]{#2}%
}

\usepackage{hyperref}

\usepackage{orcidlink}

\begin{document}

\title{Beyond Aesthetics: Quantifying Information Loss in Turbid Scenes} 

\titlerunning{PCD}

\author{Vasiliki Ismiroglou\inst{1,2}\orcidlink{0009-0009-8428-1113} \and
Stefan H. Bengtson\inst{1,2}\orcidlink{0000-0001-9306-7658} \and
Tasos Benos\inst{1}\orcidlink{0009-0004-4278-2191} \and
Thomas B. Moeslund\inst{1,2}\orcidlink{0000-0001-7584-5209}\and
Malte Pedersen\inst{1,2}\orcidlink{0000-0002-2941-9150}}

\authorrunning{V. Ismiroglou et al.}

\institute{Visual Analysis and Perception Laboratory, Aalborg University, Denmark\and
Pioneer Centre for Artificial Intelligence, Denmark}

\maketitle

\begin{abstract}
Visibility in underwater environments degrades rapidly under turbid conditions, yet the effects on computer-vision models remain unclear.
This issue is compounded by reliance on synthetic turbidity datasets, which may misrepresent real‑world information loss.
To address this gap, we introduce the Turbid Underwater Baseline (TUB) dataset, comprising 1,320 images captured under extreme turbidity and over 16,000 high-confidence ground-truth segmentation masks.
We additionally propose PCD, a metric derived from phase congruency maps that is invariant to contrast and aims to capture the loss of structural information in real turbidity. We show that PCD correlates strongly with the performance of instance segmentation models on both real and synthetic turbid images, whereas common metrics in the field show weak to no correlation at all.
The dataset and relevant code can be found on the project page: \href{https://vap.aau.dk/pcd}{https://vap.aau.dk/pcd}

  \keywords{Underwater \and Turbidity \and Metric \and Phase congruency \and Synthetic data}
\end{abstract}

\section{Introduction}\label{sec:intro}
\begin{figure}[tb]
    \centering
    \includegraphics[width=1\linewidth]{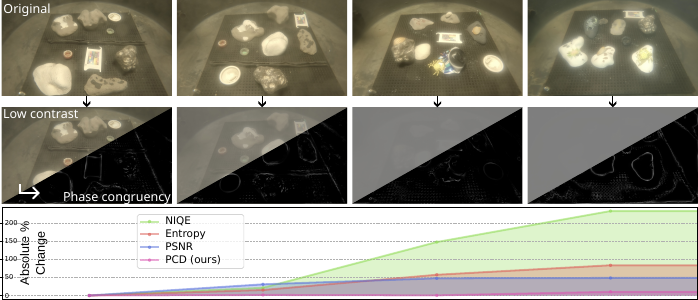}
    \caption{Four clear images from the TUB dataset are shown with progressively reduced contrast. The proposed PCD metric indicates little change in information while alternative metrics show a substantial drop in image quality. The corresponding phase congruency maps, computed from the low contrast images, highlight that a considerable amount of structural information is retained.}
    \label{sec:intro:fig:fig1}
\end{figure}
Vision systems are increasingly used for critical infrastructure monitoring, biodiversity estimation, and navigation in diverse underwater environments \cite{gonzalez-sabbagh_survey_2023, naveen_advancements_2025, jiao_vision-based_2024, pedersen_detection_2019, alvarez-tunon_underwater_2018}. These tasks often involve complex and low-visibility scenes, where light attenuation and scattering may dramatically reduce the perceptual quality of images. In such conditions, the performance of downstream computer vision models may decline drastically, which is typically attributed to perceptual degradation \cite{hu_overview_2022, yuan_survey_2022, xu_systematic_2023}. However, the reasoning for and level of the performance drop is not fully understood, as there currently exist no thoroughly annotated real-world image datasets covering the full range of underwater turbid conditions.

Turbidity refers to the cloudiness of water due to suspended particles and is a complex optical phenomenon with distinct visual effects. As light travels from the environment to the camera through the turbid medium, it undergoes absorption and scattering \cite{mobley_light_1994}. Absorption is strongly wavelength‑dependent, leading to darker, spectrally distorted images \cite{akkaynak_revised_2018, solonenko_inherent_2015}. Scattering is a more complex process in which particles redirect light at angles determined by their size and shape. As a result, peripheral light enters the camera in the form of backscatter, and light originating from the scene is either deflected away or spread across a wider area of the sensor \cite{mcglamery_computer_1980, jaffe_computer_1990}. Because turbid mediums are rarely homogeneous, these effects are further convoluted due to the varying local density of sediment. The combined effect is a characteristic visual degradation: color distortion from selective absorption, reduced contrast due to the milky backscattering veil, and in highly scattering conditions \textit{structural degradations}, in the form of blurred or otherwise corrupted object boundaries.

Owing to the scarcity of available turbid data, a substantial body of recent research has shifted toward the generation of synthetic images by modeling this light-medium interaction \cite{desai_rsuigm_2024, liu_model-based_2022, wang_uwgan_2019, fathy_submergestylegan_2023}. Nevertheless, such works have predominantly concentrated on perceptual image enhancement applications, replicating only a subset of common visual degradations such as color shifts and contrast reduction. Without a framework to quantify the information loss that occurs in high-scattering environments, it is an open question whether these synthetic effects share any meaningful similarities to the structural degradations of real turbidity. If they do not, synthetic data may contribute little to training detection models and may even introduce harmful biases. To address these problems, we collect a new dataset designed to investigate the impact of turbidity on downstream computer vision tasks and propose a novel metric that can quantify this effect on real and synthetic data. 

We present Phase-Congruency Delentropy (PCD), a metric of delentropy \cite{larkin_reflections_2016} using phase‑congruency maps \cite{kovesi_phase_2000}. Phase congruency is invariant to color shifts and contrast loss, and instead captures the structural information of the image, an aspect largely overlooked in existing work on underwater image quality evaluation. \cref{sec:intro:fig:fig1} highlights how loss of contrast can result in a decrease in perceptual quality while underlying structural information remains. In order to validate the metric, we introduce the first, \textit{non-synthetic}, instance segmentation dataset featuring varying levels of turbidity, ranging from clear water to fully opaque conditions. We analyze how instance segmentation models behave under real turbidity and under commonly used synthetic degradations, demonstrating that PCD exhibits a strong correlation with downstream model performance. To summarize, our contributions are:
\begin{itemize}
    \item The Turbid Underwater Baseline (TUB) dataset, with images systematically captured from multiple viewpoints and under varying levels of turbidity. It is the first publicly available dataset offering high‑confidence ground‑truth annotations in extremely turbid conditions, making it \textit{suitable for data-driven tasks}.
    \item PCD: A novel metric for model-centric assessment of turbid scenes, demonstrating \textit{superior correlation} with model performance, across both real and synthetic turbidity conditions compared with existing alternatives.
    \item We report baseline results for state‑of‑the‑art instance segmentation models trained and evaluated across varying levels of real turbidity. They represent the first results to directly assess how turbidity affects learning‑based models.
    \item We demonstrate that poor perceptual quality does not equate to poor task utility, whereas a reduction in structural information may. 
\end{itemize}

\section{Related Work}\label{sec:rwork}

\subsubsection{Datasets with Real Turbidity}
Publicly available datasets that capture a broad spectrum of real turbidity conditions remain limited and are often curated for image enhancement tasks \cite{liu_real-world_2019, islam_fast_2020, li_underwater_2020, sun_turbid_2024}. They therefore lack the annotations required to evaluate other downstream tasks, such as object detection or semantic segmentation.
More critically, these datasets contain few, if any, examples of severe turbidity. The TURBID dataset \cite{duarte_dataset_2016} offers a wider range of induced turbidity levels, yet it encompasses only a small number of scenes, restricting its applicability for data-driven tasks as it contains less than 100 images in total.

When examining detection‑centric datasets, the RUOD \cite{fu_rethinking_2023}, UDD \cite{liu_new_2022}, and RUIE \cite{liu_real-world_2019} datasets contain scenes that are commonly regarded as turbid. However, although the images exhibit strong color distortions, the targets themselves generally remain clear and well‑defined. FishInTurbidWaters \cite{jahanbakht_semi-supervised_2023} includes fish imagery under more severe turbidity, but its annotations are sparse, providing only binary presence–absence labels. The Brackish dataset \cite{pedersen_detection_2019} offers bounding‑box annotations in high‑turbidity conditions, yet the range of environmental variability is relatively narrow.
A key limitation of these datasets is that they do not record information about the environmental conditions, making it difficult to relate model performance to turbidity.

\subsubsection{Synthetic Turbidity Generation}

In response to the scarcity of varied real‑world data, synthetic turbidity generation has become a common strategy. Synthetic turbid datasets generally fall into two categories: learning‑based generative models \cite{fabbri_enhancing_2018, sun_turbid_2024, wang_uwgan_2019, li_synthesis_2018}, and physical-model-based methods that apply numerical approximations of a light scattering model directly to 2D images \cite{hou_benchmarking_2020, desai_rsuigm_2024, akkaynak_revised_2018, ueda_underwater_2019, ismiroglou_sea-ing_nodate, liu_model-based_2022}. The latter, commonly termed as the underwater image formation model, is particularly widespread due to its computational simplicity and is often used to create training data for underwater image restoration. 

The realism of the synthesized data is typically assessed indirectly, by using them to train image enhancement models and evaluating their ability to perceptually improve real images. Because current enhancement benchmarks rarely include extreme turbidity, synthetic pipelines are not optimized to reproduce such conditions despite their theoretical flexibility. As a result, characteristics unique to high turbidity environments, such as intense blur and heavy occlusions, are either absent from available synthetic data or not evaluated appropriately.

\subsubsection{Image-Quality Metrics} 
Image-quality metrics aim to approximate perceived quality and are often used to assess underwater images.
These metrics can broadly be categorized according to the availability of reference data: full‑reference (FR) metrics, where a degraded image is compared to a corresponding high‑quality reference, and no‑reference (NR) metrics, which attempt to estimate image quality without any ground truth. 

Two of the most common full-reference metrics used in the underwater domain \cite{li_underwater_2020, raveendran_underwater_2021, islam_fast_2020, anwar_deep_2018} are PSNR and SSIM \cite{hore_image_2010}. 
While PSNR is strictly focused on quantifying noise, it is often used as a way to gauge the perceived quality of images.
The obvious limitation of such metrics is that it is incredibly challenging to acquire true reference images underwater. In most existing cases, the reference and degraded images are artificially altered versions of each other.

Common NR metrics specifically designed for the underwater domain are UCIQE \cite{yang_underwater_2015} and UIQM \cite{panetta_human-visual-system-inspired_2016}. Both metrics are based on coefficients finetuned on specific data and may diverge if applied in drastically different conditions. BRISQUE\cite{mittal_no-reference_2012}, DIIVINE \cite{moorthy_blind_2011}, and BLIINDS \cite{saad_dct_2010} are examples of general-purpose NR metrics, which have been developed by training on databases of human-rated distortions. NIQE \cite{mittal_making_2013} was proposed as truly blind as it requires no such labels.

\subsubsection{Image-Complexity Metrics} 
Image-complexity metrics attempt to quantify the complexity or information in an image and do not necessarily consider its perceived quality. Entropy and Delentropy \cite{larkin_reflections_2016} are both information-centric NR measures that have been shown to correlate with model performance on common benchmarks \cite{rahane_measures_2020}. 
Another example is VIF \cite{sheikh_image_2006}, which is a full‑reference information-centric metric based on wavelet decomposition, and SSEQ \cite{liu_no-reference_2014} that extends similar principles to a no‑reference, learning‑based framework.

Prior work has shown that improving the perceived visual quality of underwater images does not translate into better detector performance \cite{wang_is_2024, lucas_underwater_nodate}. This suggests that existing metrics do not adequately capture the information content that matters for data-driven learning in turbid conditions. As a result, these metrics are also likely unsuitable for evaluating whether synthetic turbidity effects reproduce the information loss observed in real environments.

These findings point to the need for a dedicated evaluation measure for turbid scenes. Developing such a measure requires a dataset that contains real images with varying levels of turbidity, which would allow the degradation process to be characterized and modeled. 
To address this gap, we collect the TUB dataset which we present in \cref{sec:data}.
The TUB dataset facilitates the evaluation of the novel PCD metric introduced in \cref{sec:pcd}.

\section{The Turbid Underwater (TUB) Dataset}\label{sec:data}

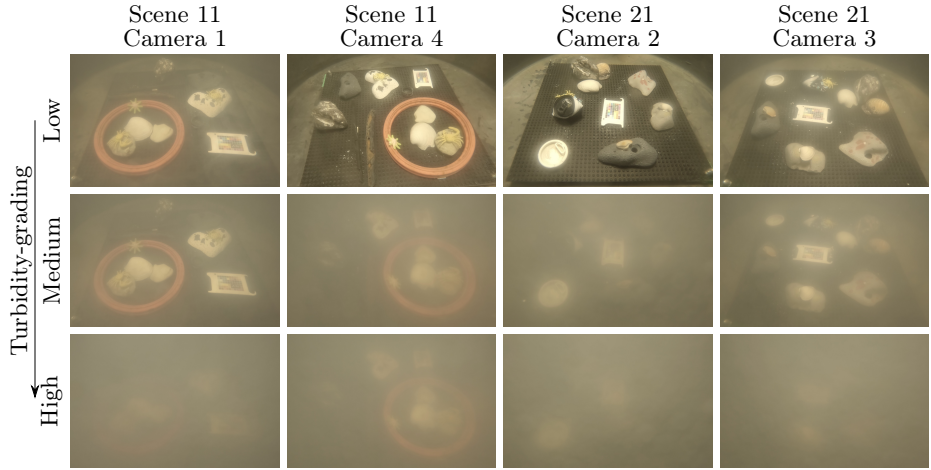
\begin{figure}[!tbp]
    \input{figures/dataset-examples}
    \caption{Example of different scenes from the TUB dataset with different levels of turbidity (grouped by NTU) and from different camera-viewpoints.}
    \label{sec:data:fig:sample_data_images} 
\end{figure}

We introduce the Turbid Underwater Baseline (TUB) dataset to address the gap in available datasets focusing on turbid underwater imagery. TUB is intended to support a wide range of research tasks involving high-turbidity environments, including but not limited to underwater image enhancement and restoration with explicit ground truth, object detection and segmentation under severe visibility degradation, and multi-view reconstruction. 
Examples of different scenes, viewpoints, and levels of turbidity in the dataset are shown in \cref{sec:data:fig:sample_data_images}.

When designing a dataset to meaningfully evaluate the impact of turbidity on vision models, other sources of uncertainty must be minimized. Cameras and objects should remain fixed across turbidity levels to ensure that comparisons reflect only the effects of degradation. The turbidity itself must be both consistent and quantitatively measured to establish a reliable ground truth. Finally, building a dataset suitable for data‑driven tasks requires not only sufficient image volume but also enough scene diversity to support robust learning.

\subsubsection{Setup}
The data‑collection setup for TUB was kept fully static, both in terms of camera placement and scene configuration. Four GoPro11 cameras were mounted around the perimeter of a cylindrical tank, and a LEGO baseplate was fixed at the bottom. Objects were attached to the baseplate using LEGO bricks and thus remained static throughout each turbidity sequence. This is the same approach as the BUCKET dataset \cite{ismiroglou_sea-ing_nodate}.

Objects were selected randomly, including natural items such as rocks and man‑made items such as trash. They were arranged in different positional configurations and object combinations, ensuring that no two scenes shared the same layout or orientations. These choices fulfilled the scene consistency requirement.
The multi‑camera setup enabled a substantial increase in image volume by capturing each scene from multiple viewpoints, naturally providing different occlusion levels without additional collection effort. This allowed more scenes to be gathered within the same timeframe. 

\subsubsection{Protocol} \label{sec:data:protocol}
Turbidity was introduced by filling the tank with water and adding measured amounts of oat milk. A total of 10 levels of turbidity were induced for every scene, with the first being clear water. 
A nephelometer was used to obtain accurate turbidity readings from three water samples at every stage.

Because turbidity levels varied across experiments, even when oat‑milk increments were measured accurately, the same nominal stage (e.g., “stage 3”) could still appear noticeably different between scenes. Therefore, nephelometer readings were used to regroup the images into three turbidity categories: low (0–10 NTU), medium (10–20 NTU), and high (20+ NTU).
This grouping serves two purposes. First, it produces a clear visual separation: objects are distinctly visible in the low group, partially degraded in the medium group, and heavily obscured in the high group. Second, the resulting distribution of images is relatively balanced across the three sets, making them suitable for training and evaluation experiments. 

In total, the dataset contains 1,320 images, with 40 images per scene (4 viewpoints × 10 turbidity levels) across 33 unique scenes. This makes it the largest dataset of its kind, providing clear‑image references while also capturing extreme turbidity conditions.

\subsubsection{Annotations}
\begin{figure}
    \centering
    \includegraphics[width=1\linewidth]{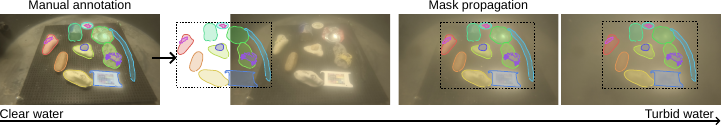}
    \caption{Annotation pipeline. Masks are delineated in the clear images and extended to all turbidity levels due to the static nature of the setup. }
    \label{sec:data:fig:ann_pipeline}
\end{figure}

Manual annotations were created for all visible objects, resulting in over 16,000 instance‑level segmentation masks. A single class label was used during annotation, with no distinction made between different object types. All scenes were annotated under clear (non‑turbid) conditions using the images captured prior to adding oat milk. These annotations were then propagated to all corresponding turbid images, as the physical arrangement of each scene remained fixed across turbidity levels, as illustrated in \cref{sec:data:fig:ann_pipeline}.

This approach makes TUB the first underwater dataset to provide high confidence ground‑truth labels even under severe turbidity, where human annotators would otherwise be unable to accurately delineate object boundaries.

\section{Delentropy of Phase Congruency Maps}\label{sec:pcd}

While a vast range of image quality and complexity metrics exist and have been used in the underwater domain, the majority of them aim to align with human perception. However, deep-learning architectures have the theoretical capacity to extract information from images regardless of whether a human can perceive that same information.

Prior work has established that the principal structural information in an image resides in the phase component of its Fourier transform \cite{oppenheim_importance_1981}. We hypothesize that this structural information is not entirely lost even when affected by attenuation and backscatter typical of underwater environments. We consider forms of discoloration and contrast distortion that still preserve shapes and region boundaries to represent a domain‑shift phenomenon and therefore be largely recoverable by deep learning models. Conversely, when severe contrast loss or wavelength‑dependent attenuation causes originally distinct neighboring regions to converge to identical pixel values, the underlying gradients may be irretrievably lost. Similar loss can occur when edges are heavily blurred or corrupted, causing boundaries to diffuse or disappear.

Delentropy already provides a framework to measure image complexity through the distribution of gradients. Originally rooted in the feature analysis of Haralick et al. \cite{haralick_textural_1973} and later described by Larkin \cite{larkin_reflections_2016} as a natural entropy measure for images, delentropy is defined as
\begin{equation}
    H(\nabla f) = H(f_x, f_y) = - \sum_{i=1}^J\sum_{i=1}^I p_{i,j}log_2 p_{i,j}
\end{equation}
where $p_{i,j}$ denotes the deldensity, the joint probability density of the gradient components $f_x, f_y$. In standard implementations, these gradients are estimated using fixed kernels. As a consequence, they are directly affected by contrast and illumination variations: regions with reduced contrast yield low gradient magnitudes, even when their underlying shapes remain discernible. Furthermore, the single-scale calculation of the gradients prevents the metric from capturing structural cues across different spatial extents.

To address these challenges, we investigate the use of phase congruency representations. Inspired by Morrone et al.’s \cite{morrone_mach_1986} observation that salient features occur where local frequency components are maximally in phase, Kovesi \cite{kovesi_phase_2000} proposed phase congruency as a low-level image invariant demonstrating robustness to illumination and contrast variations.

Phase congruency maps produce values in the range [0,1] and are computed across multiple scales. This multiscale formulation yields intensity variations that reflect edge sharpness, while still capturing broader or blurred boundaries. Visual examples of phase congruency images can be seen in \cref{sec:intro:fig:fig1}.

Kovesi’s implementation employs wavelets derived from Gabor functions and includes mechanisms for noise suppression in the estimation process.
While phase congruency can be computed in an arbitrary number of orientations for 2D images, we follow the derivation used in delentropy and compute it over two orthogonal orientations. This produces phase‑congruency‑based gradient maps. By replacing the kernel‑derived gradients used in delentropy with these phase congruency maps, we introduce a new metric that we refer to as PCD.

\section{Experiments}\label{sec:exp}

\subsection{TUB Baseline}\label{sec:exp:subsec:turb_baseline}
We first perform baseline experiments on the TUB dataset to evaluate model performance and behavior on real turbid images. We partition the data into splits at the scene level, keeping 8 scenes for validation and 25 for training. This ensures that while the same objects may appear in both splits, they never do so in the same spatial configuration or position. We further create subsets based on the measured turbidity values by dividing the images into low, medium, and high turbidity, as described earlier in \cref{sec:data:protocol}.

\begin{figure}[tb]    
    \centering
    \includegraphics[width=\linewidth]{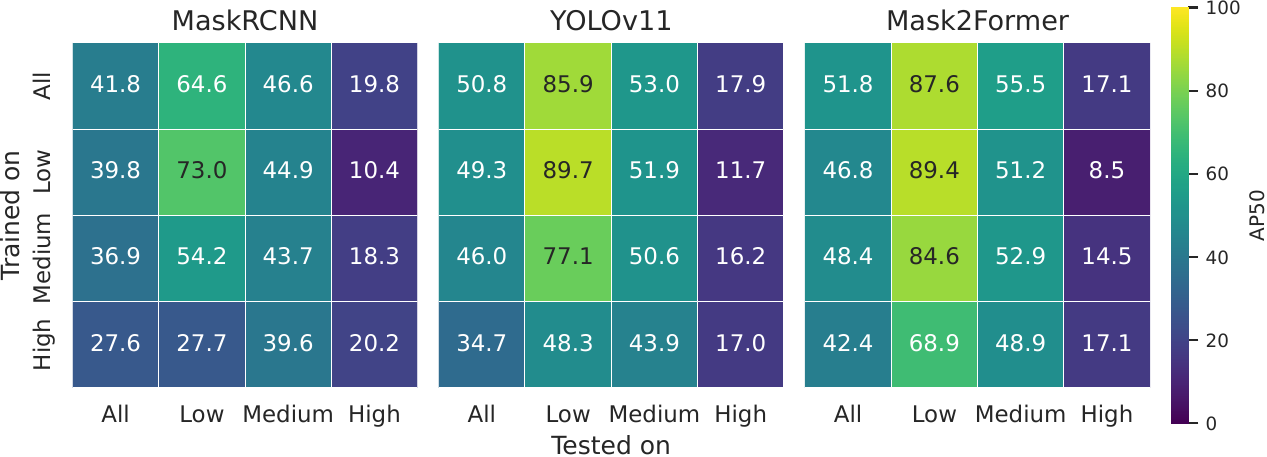}
    \caption{The performance reported as AP50 for MaskRCNN, YOLOv11, and Mask2Former when trained and tested on low, medium, and high turbidity images from the TUB dataset.}
    \label{fig:exp:bucket_benchmark}
\end{figure}

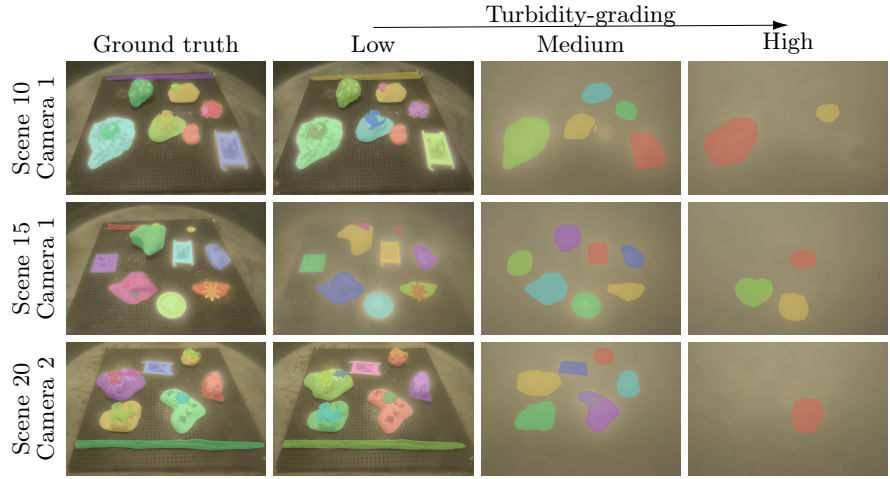
\begin{figure}[tb]
    \input{figures/predictions-examples/prediction_examples}
    \caption{Example of predicted masks from the Mask2Former model trained on the TUB dataset. All images are from the validation split. Colors are for visualization only.}
    \label{sec:exp:fig:mask_overlay}
\end{figure}

We focus on instance segmentation in the evaluation, as it is a common task within underwater computer vision.
We train and evaluate three state-of-the-art models, MaskRCNN\cite{he_mask_2018}, YOLOv11\cite{jocher_ultralytics_2023}, and Mask2Former\cite{cheng_masked-attention_2022}. These collectively cover both major architecture types for vision, i.e., convolutional neural networks and transformers. Implementation details vary between architectures due to sensitivities to learning rate and memory limitations. At the chosen training iterations, all models had converged with minor variations in losses and validation metrics. Details can be found in \cref{sec:exp:tab:settings}. The results in the form of AP50 are shown in \cref{fig:exp:bucket_benchmark}, highlighting the relationship between model performance and measured turbidity. Furthermore, we show some qualitative examples of predicted masks in \cref{sec:exp:fig:mask_overlay}.

We found the behavior across models to be very consistent despite the variations in absolute performance. Our results indicate that an increase in turbidity affects predictive capabilities even when there is no domain shift to consider. Nevertheless, all models saw improvements in the \textit{medium} and \textit{high} turbidity settings when the respective data were available during training. Extremely turbid data appear to be introducing ambiguity to the model, with performance dropping for the clean images when included. This does not affect all models equally.

\begin{table}[tb]
    \centering
    \caption{Training configuration of the different instance segmentation models.}
    \begin{tabular}{l c c c c}
    \toprule
    \textbf{Model}& Arch.               & Epochs\tablefootnote{MaskRCNN and Mask2Former were trained using the detectron2 \cite{wu2019detectron2} framework, where 6000 steps are used instead of epochs.} & LR & Batch\\\hline
      \textbf{MaskRCNN} & ResNet101-FPN \cite{he_deep_2015}&48     & $10^{-5}$& 8 \\
      \textbf{YOLOv11}   & YOLO11m-seg &50      & $10^{-3}$& 32  \\
      \textbf{Mask2Former} &  SwinB \cite{liu_swin_2021}   &48    & $10^{-4}$& 8 \\ \bottomrule
    \end{tabular}
    \label{sec:exp:tab:settings}
\end{table}

\subsection{PCD Evaluation}\label{sec:exp:subsec:metric}
We additionally aim to evaluate how well existing image‑ quality and complexity metrics, including our proposed PCD, correlate with performance in downstream tasks when using turbid data. 
For this task we include both real images and physical-model-based synthetic data. The intention is not to perform a systematic assessment of the synthetic data itself, but to examine how simplified representations of real degradation phenomena influence the behavior of these metrics.

Using the same scene-based split described in \cref{sec:exp:subsec:turb_baseline}, we take the lowest turbidity images and generate ten synthetic variants per image through two different 2D, image-based, synthetic degradation methods. The first applies the underwater image‑formation model \cite{desai_rsuigm_2024, akkaynak_revised_2018}, treating turbidity primarily as distance‑dependent contrast loss and will be named $Synth_1$ for ease of reference. The second model, $Synth_2$\cite{ismiroglou_sea-ing_nodate}, extends this approach by incorporating blur and spatially varying inhomogeneity noise.

Image formation model approaches require scene geometry and attenuation coefficients to produce images. We employ ZoeDepth \cite{bhat_zoedepth_2023} as a monocular depth estimator and randomly rescale its outputs between $0.2m$ and $4m$. We also sample attenuation coefficients from a distribution derived from the Jerlov water types \cite{solonenko_inherent_2015}. This setup reflects typical scenarios where scene geometry and the exact model parameters are unknown. The synthetic images retain the same mask annotations as their real counterparts, and both sets are merged to form an expanded dataset. In total, this triples the volume of the images.

Similar to the turbidity baseline, we train MaskRCNN, YOLOv11, and Mask2-Former on the combined dataset and report AP50 separately per image. We also compute a broad set of quality and information‑content metrics. Shannon entropy and Delentropy are included because they directly relate to the proposed PCD measure as indicators of information and structural complexity. In addition, we compute three no‑reference quality metrics: NIQE, which is designed for general natural‑image statistics, and UIQM and UCIQE, which are tailored to underwater imagery. Finally, we measure SSIM and PSNR using the lowest‑turbidity images as a reference to quantify structural and photometric deviations. We do not fine-tune the statistics or coefficients of any of the metrics on our data.

\begin{figure}[tb]
    \centering
    \includegraphics[width=\linewidth]{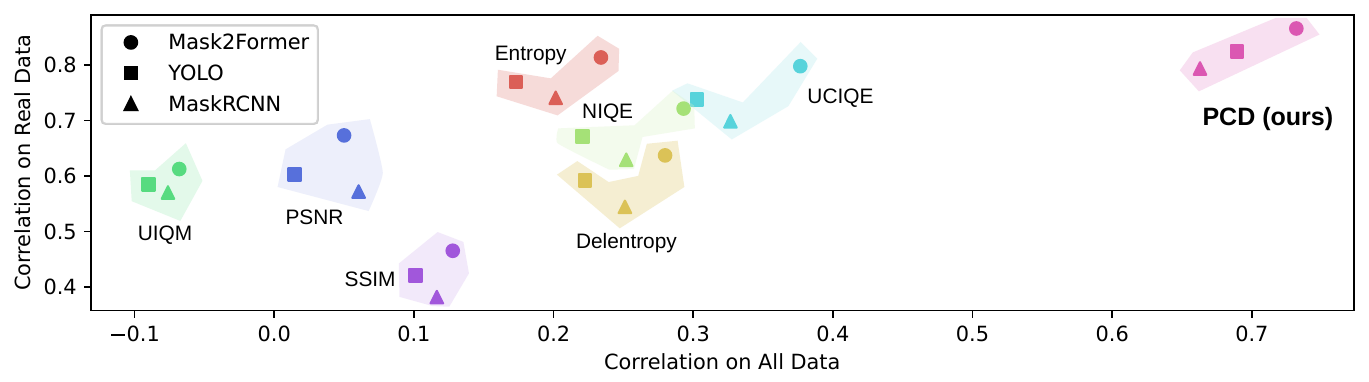}
    \caption{Metric correlation with model performance. The area highlights do not represent any measurable quantity, but are simply used to assist visualization.}
    \label{sec:exp:fig:corr_statistics}
\end{figure}

\begin{figure}[p]
    \centering
    \includegraphics[width=\linewidth]{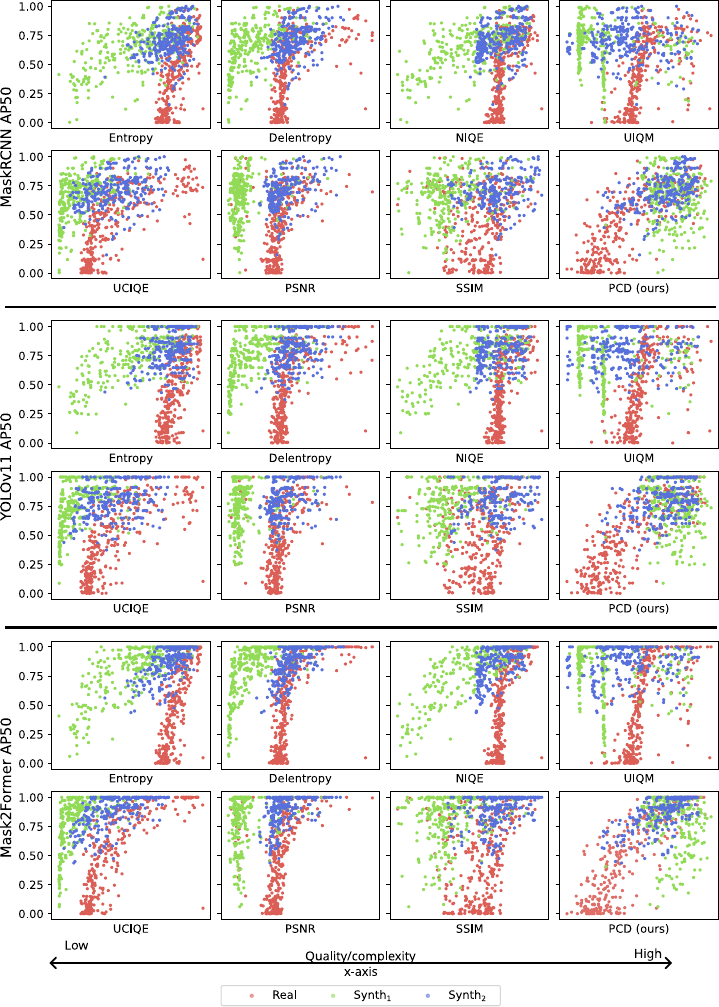}
    \caption{Scatterplots highlighting correlation between model performance and common image quality metrics for real and synthetic data. The models listed from top to bottom are MaskRCNN, YOLOv11, and Mask2Former. NIQE is inverted for visual consistency.}
    \label{sec:exp:fig:metrics_vs_maskrcnn}
\end{figure}

The correlation of different metrics with model performance is summarized in \cref{sec:exp:fig:corr_statistics} while the full scatterplots are shown in \cref{sec:exp:fig:metrics_vs_maskrcnn}. We find that many of the metrics exhibit weak correlation even on real images. Turbid scenes often collapse into narrow metric ranges despite clear variation in object visibility as reflected by model performance. We also observe a consistent pattern across models: MaskRCNN shows the lowest correlations with all metrics, followed by YOLOv11 and Mask2Former, a hierarchy that aligns with their predictive capabilities.

When evaluating image quality on synthetic data, all established metrics fail to reflect the actual degree of degradation, as shown in \cref{sec:exp:fig:metrics_vs_maskrcnn}. A recurring trend is that the synthetic images, particularly the ones of $Synth_1$, are scored as substantially lower quality than real turbid images, despite the models maintaining high‑precision predictions on many of them. On the contrary, PCD produces higher scores, similar to the ones of clear images, for the majority of $Synth_1$. While this results in some outliers, there is also a meaningful portion of overlap with the ground-truth distribution.

\begin{figure}[tb]
    \input{figures/predictions-examples/synthetic-prediction_examples}
    \caption{Example of predicted masks on synthetic images.
    First row: ground truth masks on non-turbid images from the TUB dataset. 
    Second and third row: predicted masks on images with synthetic turbidity generated using $Synth_1$ and $Synth_2$, respectively.
    Predictions are from Mask2Former, and all images are from the validation split. Colors are for visualization only.}
    \label{sec:exp:fig:synth_mask_overlay}
\end{figure}
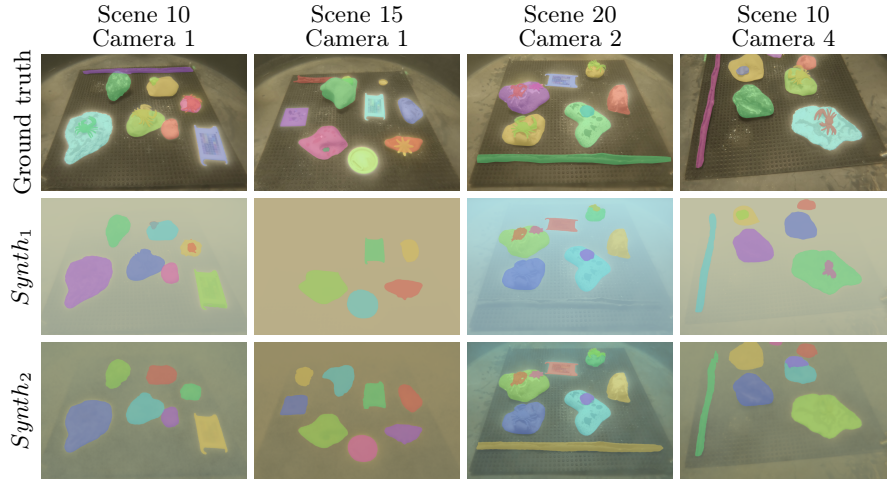

A visual example to explain this behavior can be seen in \cref{sec:intro:fig:fig1,sec:exp:fig:synth_mask_overlay}. When contrast is reduced without any blur or noise, gradients remain sharp. The majority of metrics already consider that a drop in quality, but model performance is not affected drastically. After a point, objects further from the camera disappear completely. Gradients are still sharp and indicative of a clear image, but detector performance drops as they are evaluated on objects that are fully occluded rather than degraded.

Overall, PCD shows the highest correlation with model performance, indicating that gradient degradation is a better indicator of image quality in high turbidity settings than contrast, discolorations, or pixel statistics. This is further supported by the differences in the synthetic data when gradient-degrading effects are considered.

\subsubsection{Phase-congruency hyperparameters}
PCD does not rely on training data or statistical priors, but its phase‑congruency computation includes hyperparameters that can be tuned. Kovesi \cite{kovesi_phase_2000} noted that the impact of the hyperparameters is limited and illustrated his point with a few qualitative examples. Nevertheless, we believe that in images dominated by noise and diffuse gradients, such as those typical of turbid environments, the wavelet characteristics over which phase congruency is calculated are critical. Using only a few scales emphasizes high‑frequency structure, whereas increasing the number of scales progressively incorporates lower‑frequency information and diminishes the relative weight of fine‑scale content. 

\begin{figure}[tb]
    \centering
    \includegraphics[width=1\linewidth]{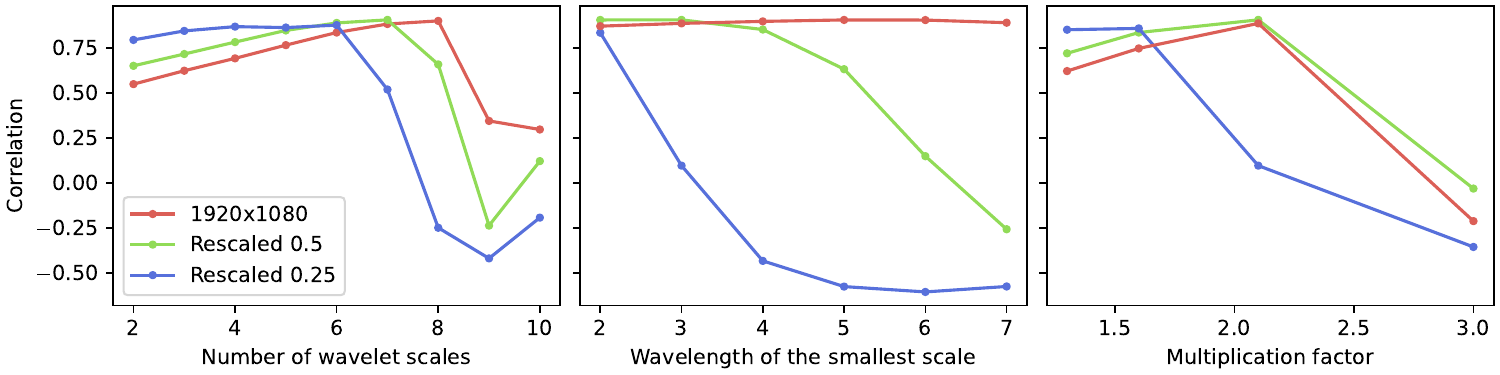}
    \caption{Variation of PCD correlation to Mask2Former AP50 as the wavelet characteristic over which phase congruency is calculated change.}
    \label{sec:exp:fig:nscales}
\end{figure}

To assess this sensitivity, we analyze PCD's correlation with Mask2Former across a range of image‑rescaling factors and wavelet‑scale configurations. This allows us to evaluate how robust the metric remains under different parameter choices. The results are presented in \cref{sec:exp:fig:nscales}. For the experiments reported in the main text, we downsampled the images by a factor of 0.5 and used seven wavelet scales with a minimum wavelength of 3 and a multiplication factor of 2.1. However, our analysis indicates that six wavelet scales with a minimum wavelength of 2 provide more stable behavior overall.

\section{Discussion}\label{sec:discussion}
Baseline turbidity experiments show that clear images are essential, as models trained solely on turbid data can exhibit reduced performance even in the absence of a domain gap. This likely reflects the role of clear images in enabling stronger feature representations, which have the capacity to generalize to novel viewpoints and conditions. Training on turbid data facilitates this generalization to degraded environments, often at the cost of reduced performance in clear images. This trade‑off underscores the need for strategies that introduce degraded imagery into training pipelines without compromising performance across the full range of operating conditions.

The results also expose a practical challenge: incorporating highly turbid images into training becomes difficult when the turbidity level prevents reliable human annotation. While synthetic data can help tackle this problem, our findings show that commonly used synthetic degradations are not interpreted by existing image‑quality metrics in the same way as real turbidity. Through our experiments, we demonstrate that structural degradations are more critical to detection models than perceptual ones, making PCD a better-suited metric at quantifying the effects of both real and synthetic turbidity on downstream tasks.

\subsubsection{Caveats} Non‑reference metrics, particularly those intended to approximate information content, are inevitably influenced by the underlying scene structure, independent of degradation. This makes comparisons unreliable when images differ substantially in inherent detail, such as between a featureless, clear‑water region and a highly textured scene under strong turbidity. In our case, the dataset was collected under controlled conditions, and all scenes contain comparable levels of structural detail. This consistency makes relative comparisons between metrics and degradation methods meaningful.

The dataset design also enables the isolation of error sources. Because the model’s predictions on the corresponding clear images are available, they serve as a reference to distinguish errors caused specifically by turbidity from those arising due to other factors, such as out‑of‑distribution content or occlusions. Without these conditions, evaluating such correlations would be far less straightforward, as it would be difficult to attribute failure cases to degradation rather than unrelated scene variations.

\section{Conclusion}
In this work, we introduce and evaluate PCD, a novel metric that focuses on structural degradation caused by turbidity rather than perceptual changes. Our experiments demonstrate that PCD exhibits the strongest correlation with performance on instance segmentation models across both real and synthetic turbidity when compared with existing alternatives. We thus highlight its potential for assessing turbid data information content as well as evaluating the realism of synthetic turbidity generation. We additionally present TUB, the first publicly-available instance segmentation dataset designed for extreme turbidity conditions and suitable for data‑driven learning tasks such as the PCD experiments. This dataset addresses a major gap in the existing literature and provides an essential resource for advancing vision research in turbid underwater environments.

\section*{Acknowledgements}
This work was funded by the Pioneer Centre for AI, DNRF grant number P1. We further gratefully acknowledge DTU Aqua for providing access to their nephelometer for TUB measurements.

%
%

\bibliographystyle{splncs04}
\bibliography{main}
\end{document}

%% file: figures/dataset-examples.tex
\newcommand{\rh}{1.75cm} 

\begin{tikzpicture}[
    every node/.style={anchor=center, inner sep=0, outer sep=0},
    row sep=0pt,
    column sep=3pt
]

\matrix (m) [
    matrix of nodes,
    nodes={anchor=center},
    row sep=3pt,
    column sep=3pt
]{

      {} 
    & \shortstack{Scene 11\\Camera 1}
    & \shortstack{Scene 11\\Camera 4}
    & \shortstack{Scene 21\\Camera 2}
    & \shortstack{Scene 21\\Camera 3}
    \\

    \parbox[c][\rh][c]{0.3cm}{\centering \rotatebox{90}{Low}} &
    \cropleft[height=\rh]{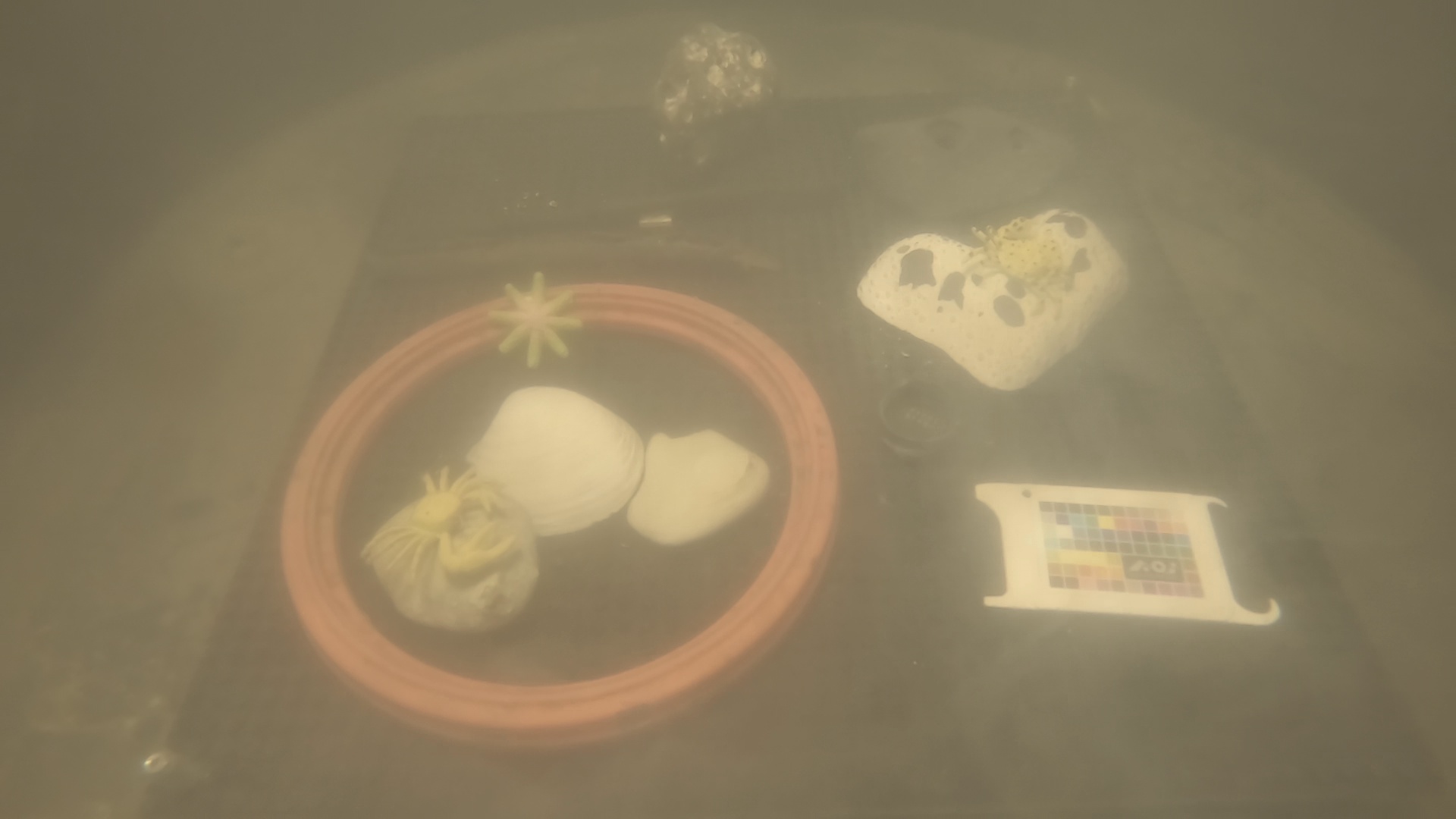} &
    \cropleft[height=\rh]{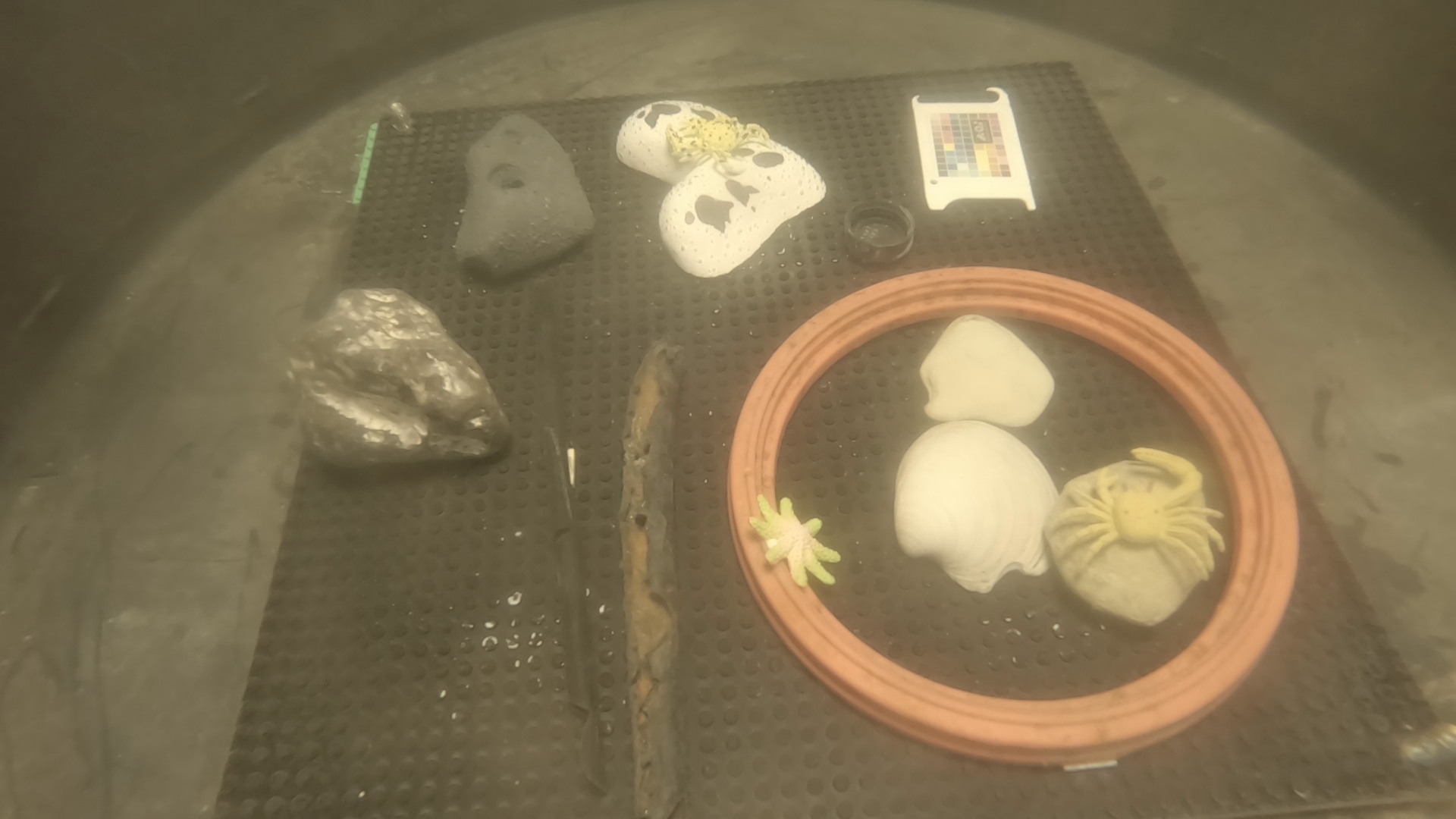} &
    \cropleft[height=\rh]{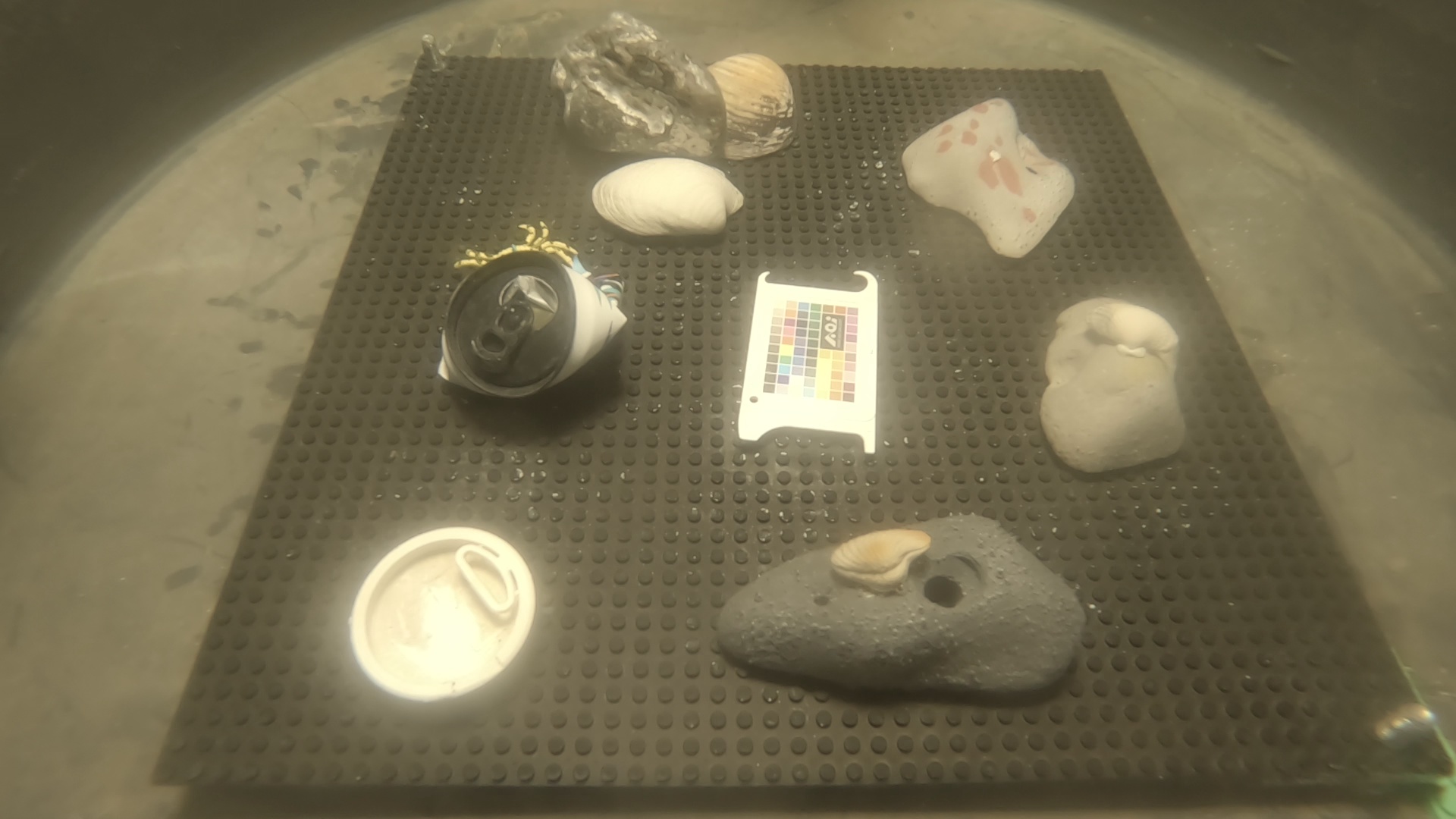} &
    \cropleft[height=\rh]{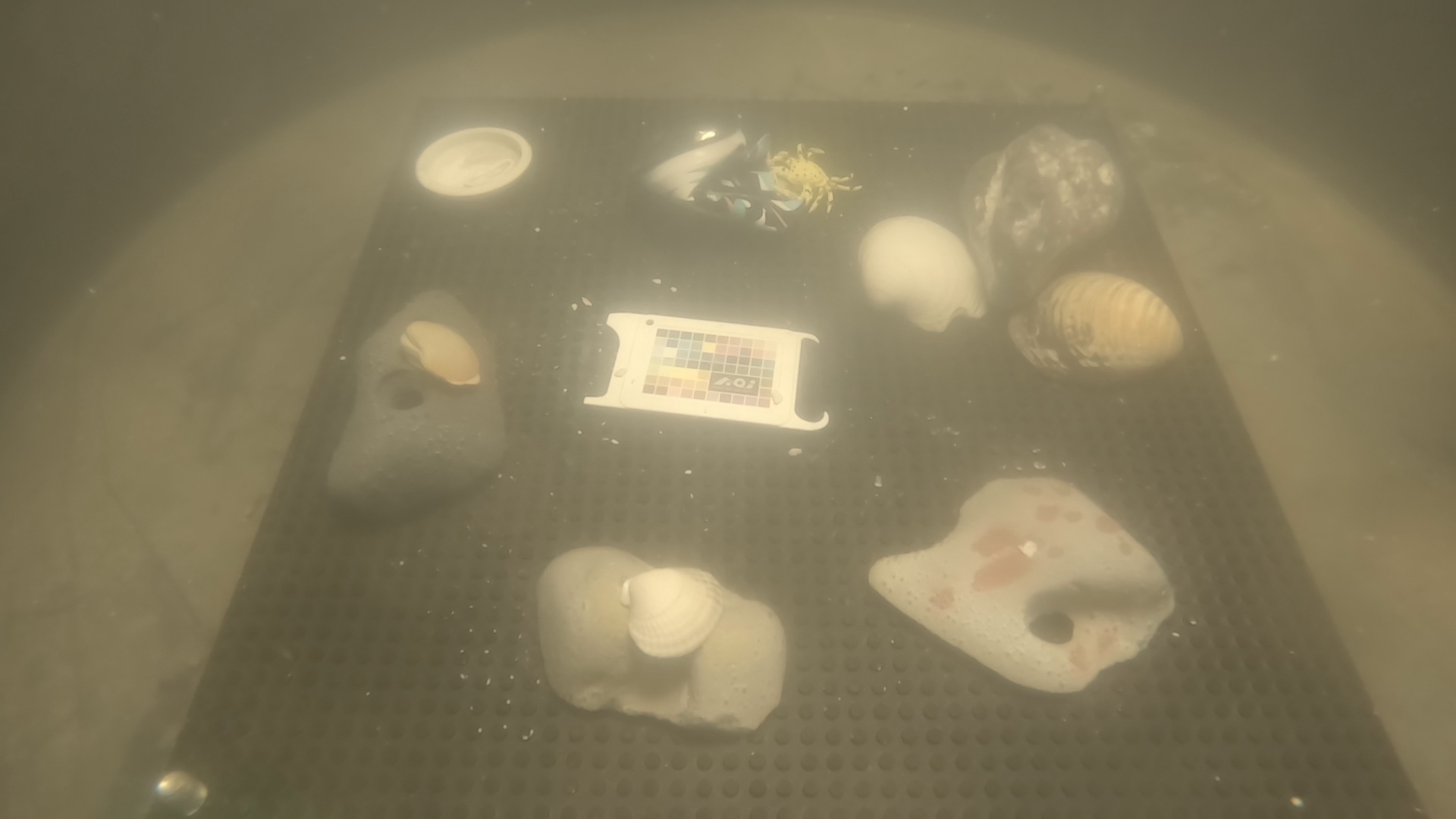}
    \\

    \parbox[c][\rh][c]{0.3cm}{\centering \rotatebox{90}{Medium}} &
    \cropleft[height=\rh]{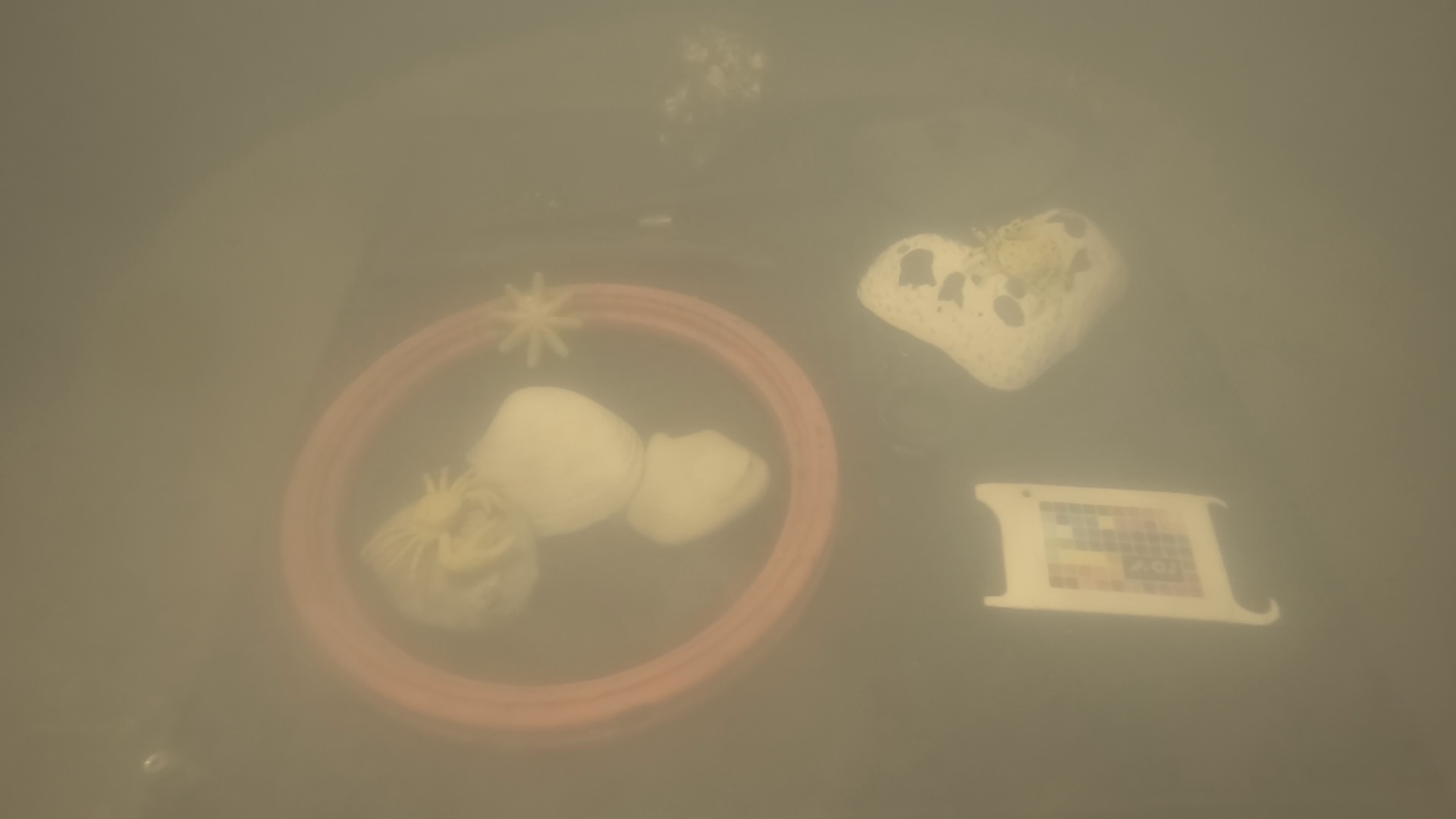} &
    \cropleft[height=\rh]{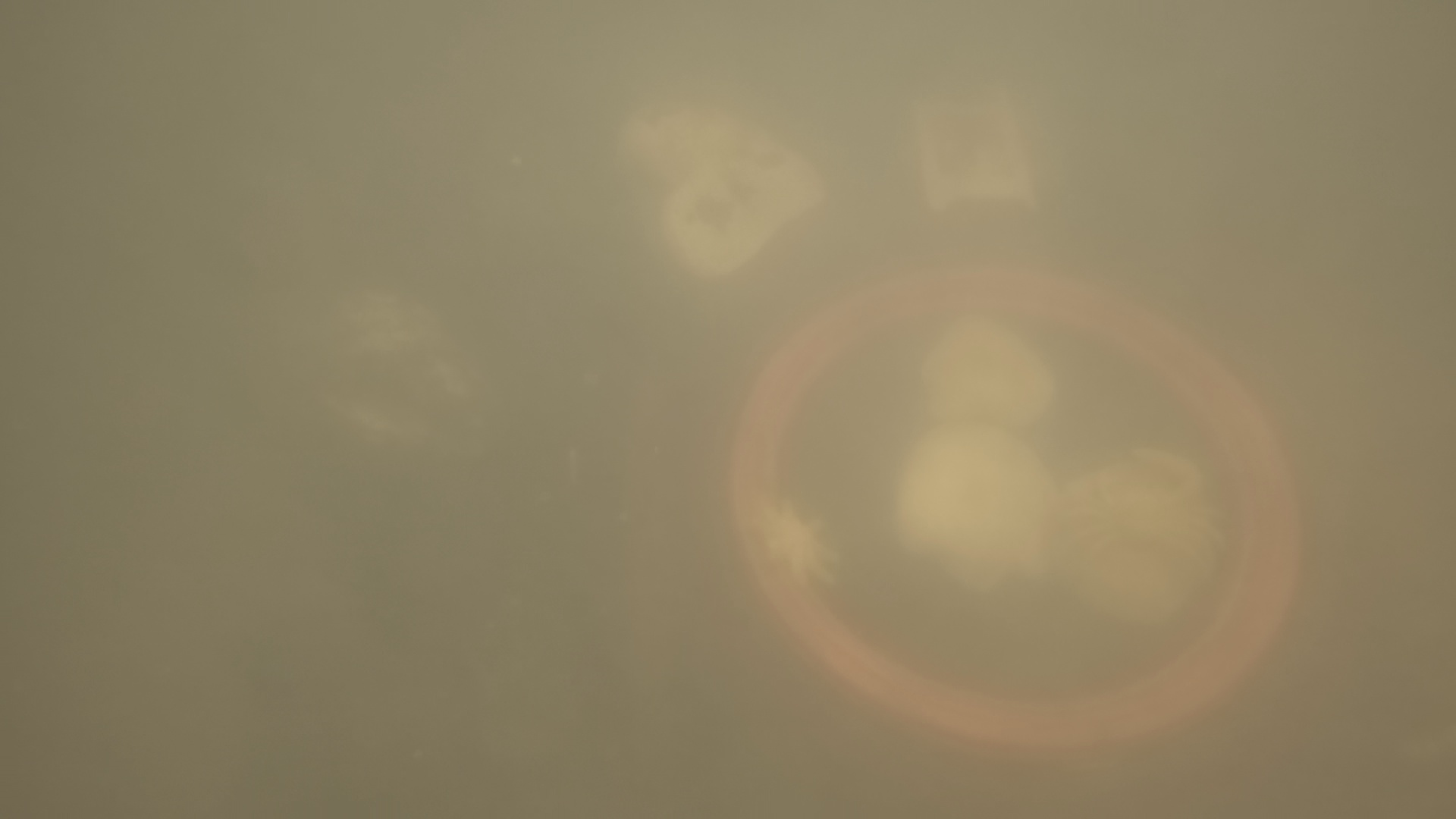} &
    \cropleft[height=\rh]{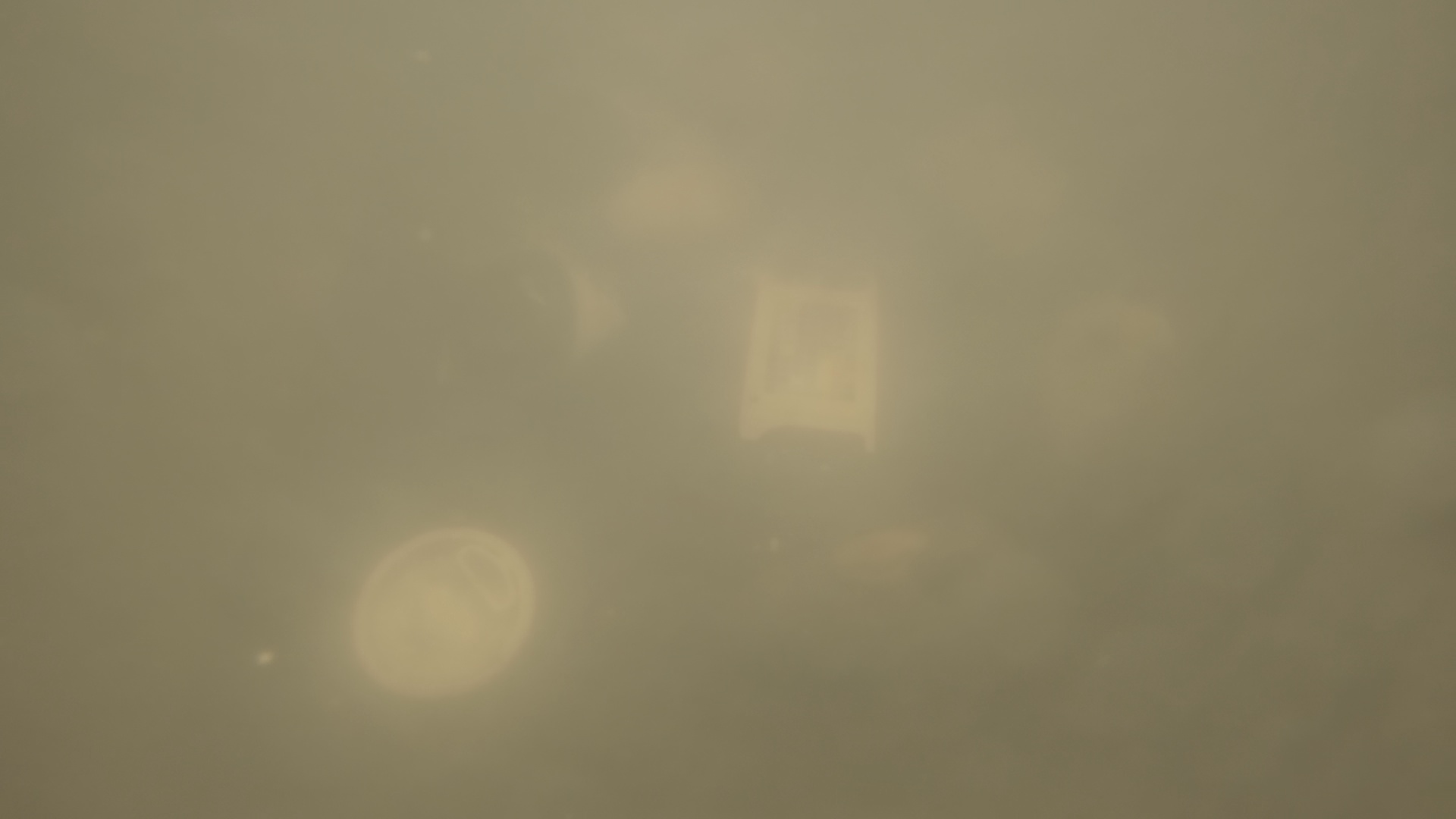} &
    \cropleft[height=\rh]{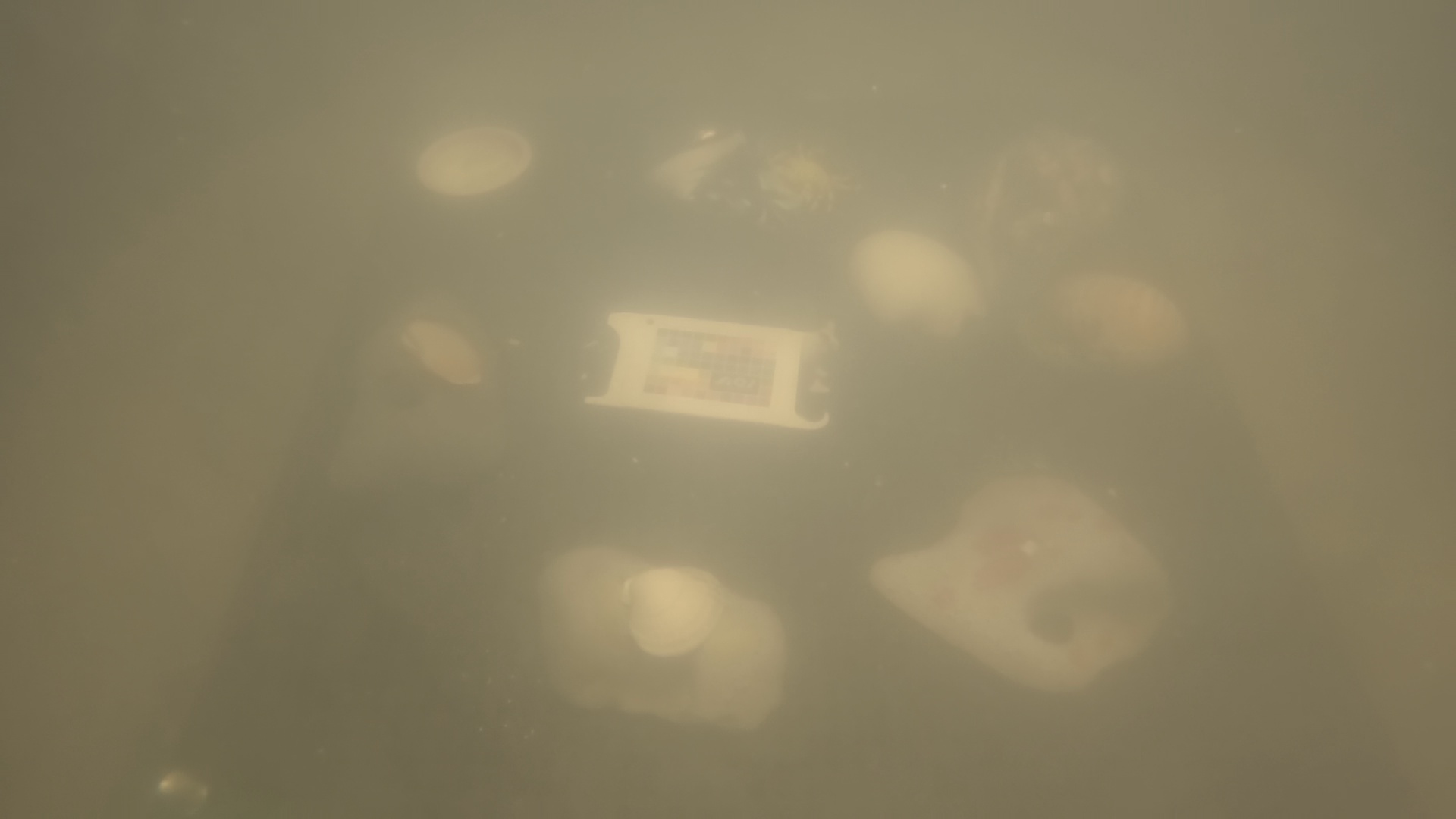}
    \\

    \parbox[c][\rh][c]{0.3cm}{\centering \rotatebox{90}{High}} &
    \cropleft[height=\rh]{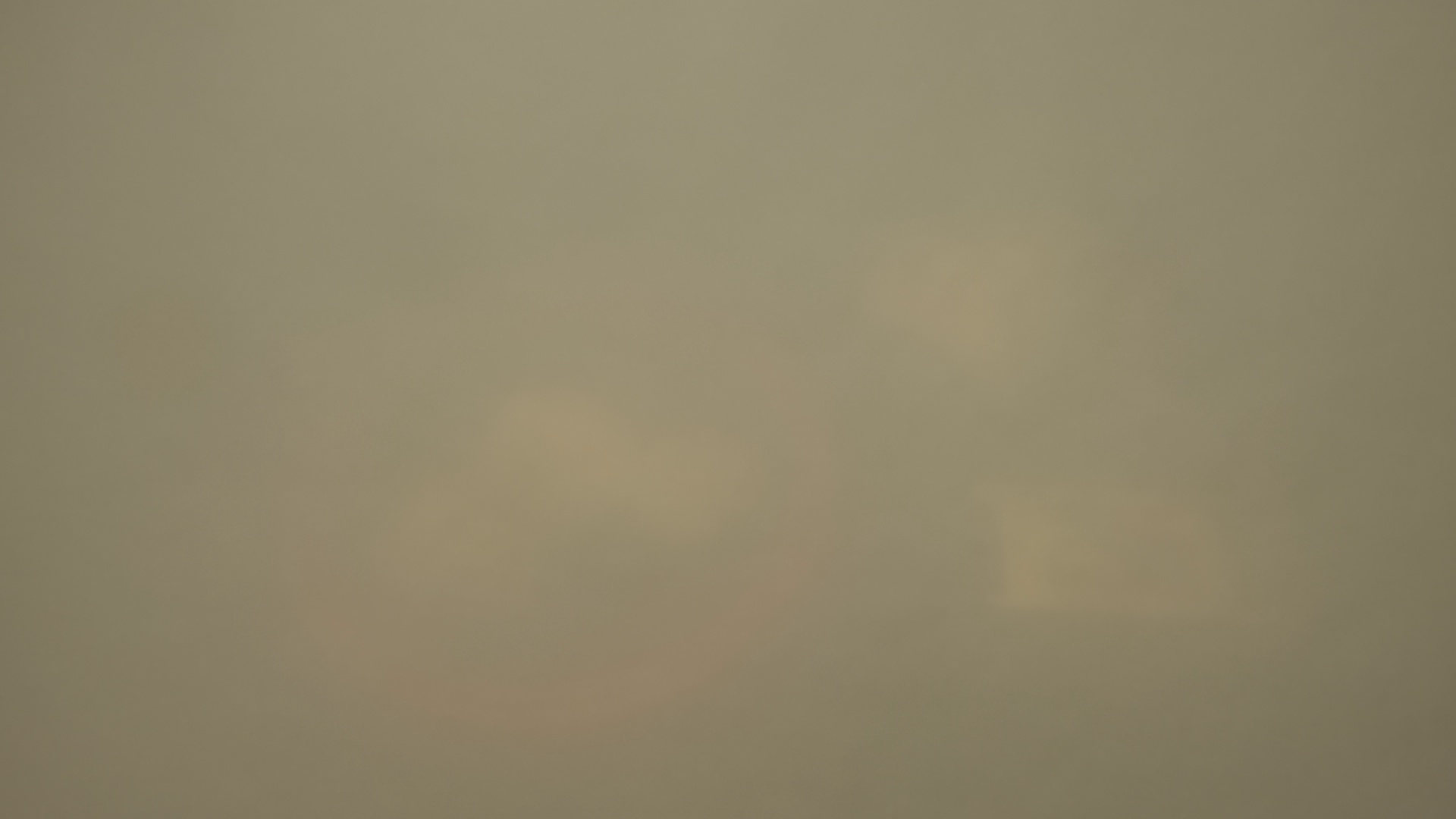} &
    \cropleft[height=\rh]{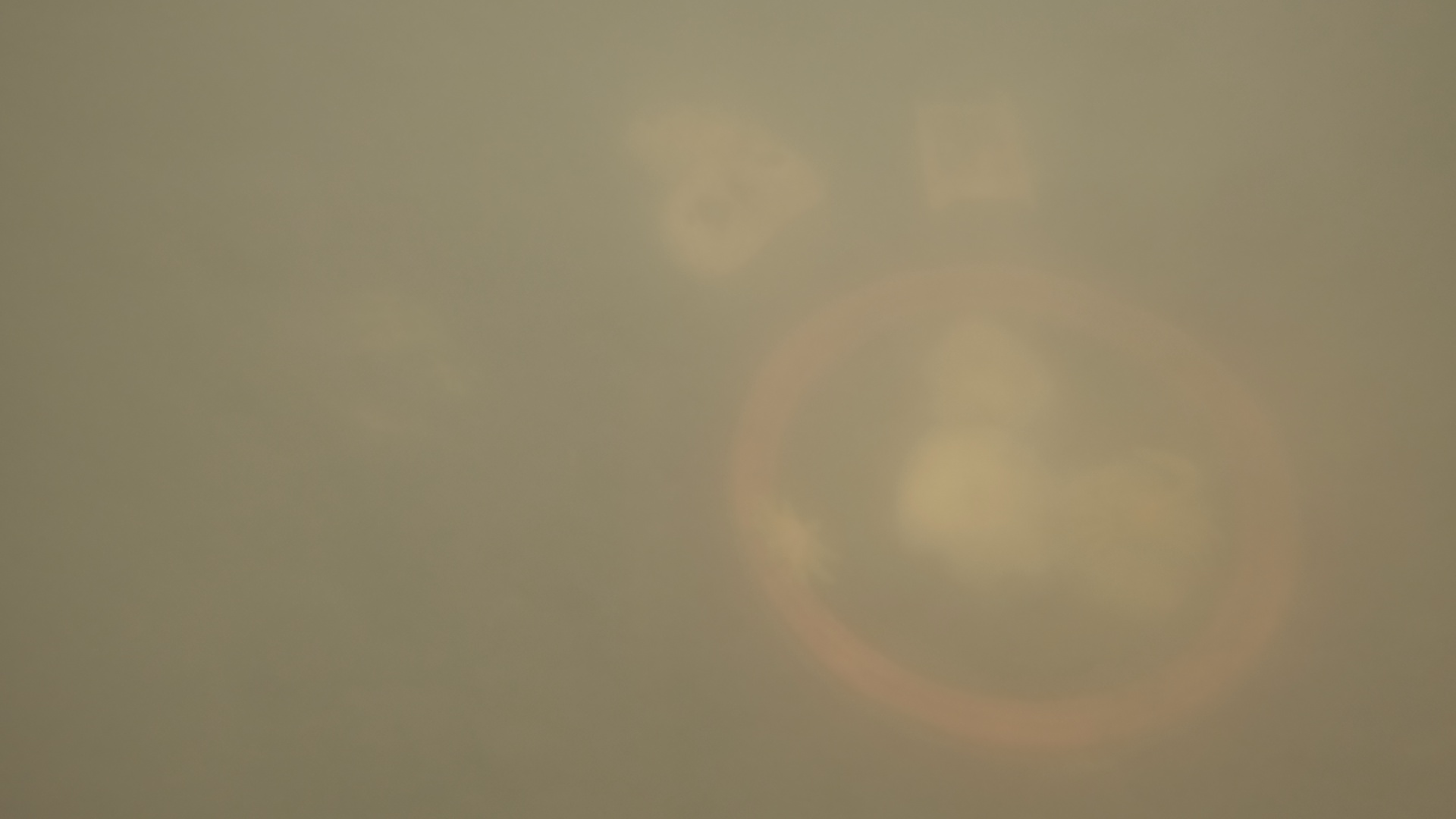} &
    \cropleft[height=\rh]{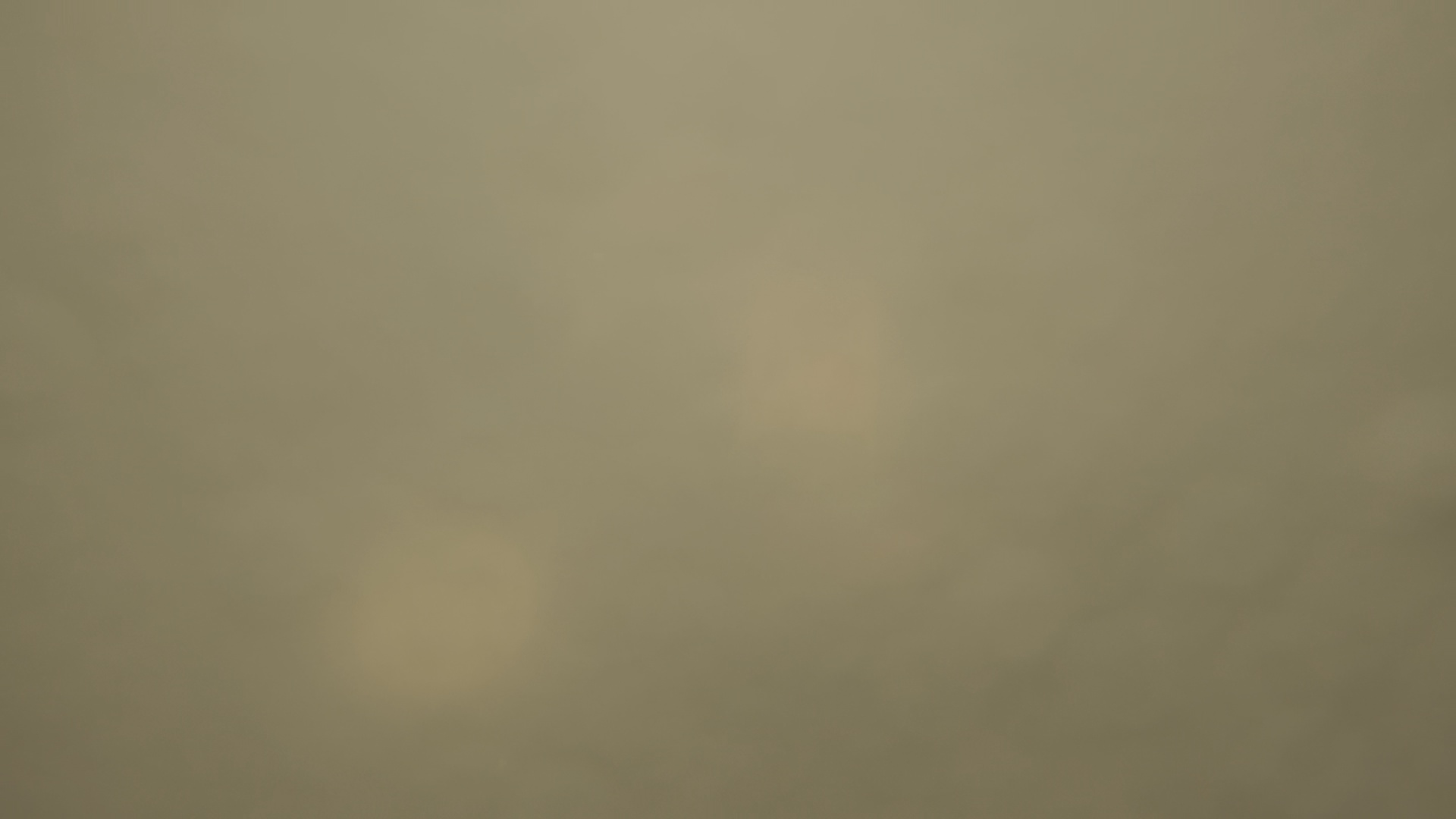} &
    \cropleft[height=\rh]{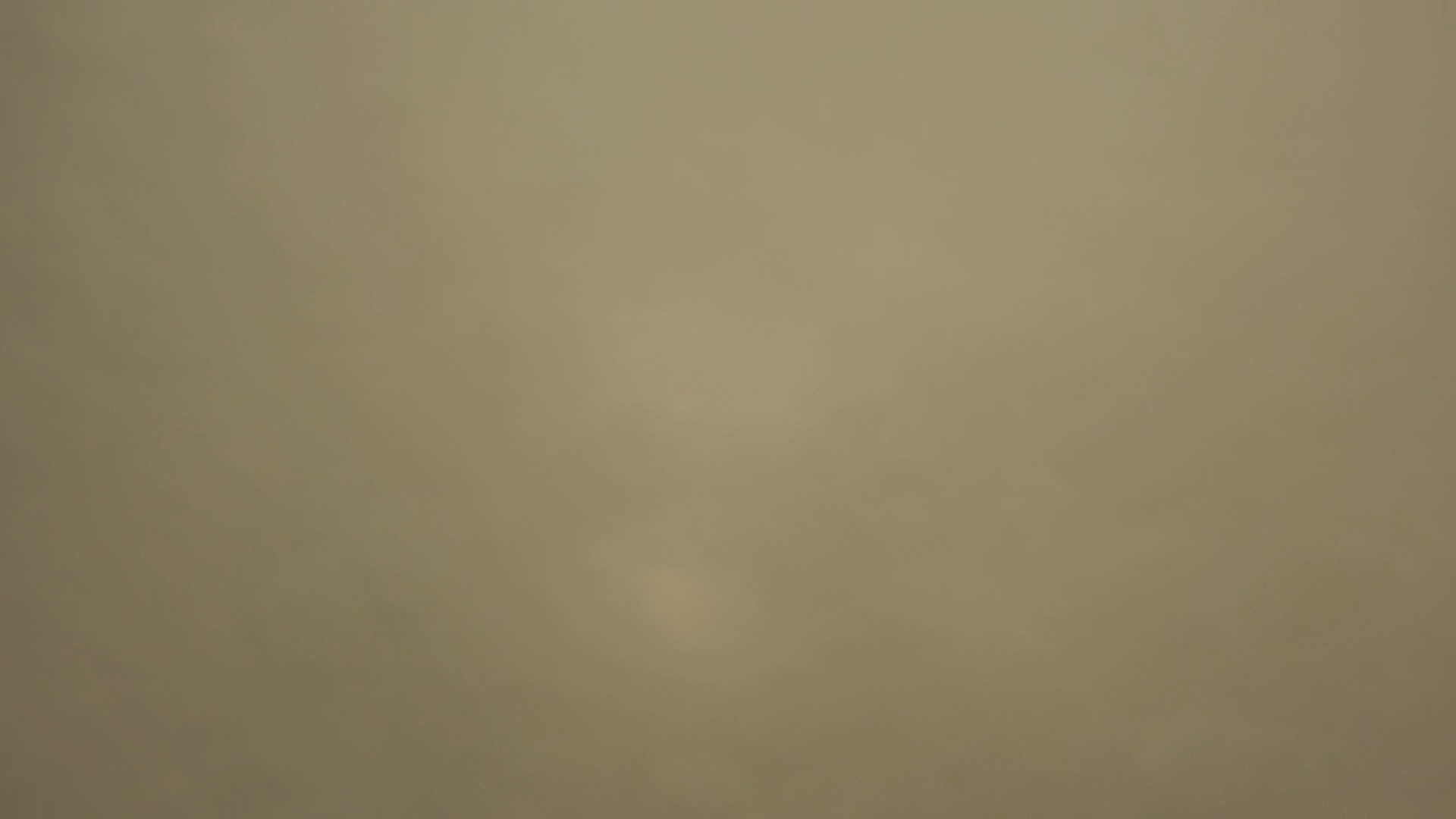}
    \\
};

\node[rotate=90, anchor=center] at ($ (m-2-1.center)!0.5!(m-4-1.center) + (-0.4cm,0) $) {\small Turbidity-grading};

\coordinate (tganchor) at ($ (m-2-1.center)!0.5!(m-4-1.center) + (-0.4cm,0) $);
\coordinate (tgarrowx) at ($ (tganchor) + (0.18cm,0) $);

\draw[line width=0.3pt,-{Stealth[length=1.8mm, width=1.1mm]},shorten >= 1pt] (tgarrowx |- m-2-1.center) -- (tgarrowx |- m-4-1.center);

\end{tikzpicture}

%% file: figures/predictions-examples/prediction_examples.tex
\centering
\newcommand{\rh}{1.75cm}

\begin{tikzpicture}[
    every node/.style={anchor=center, inner sep=0, outer sep=0},
    row sep=0pt,
    column sep=3pt
]

\matrix (m) [
    matrix of nodes,
    nodes={anchor=center},
    row sep=3pt,
    column sep=3pt
]{

      {}
    & Ground truth
    & Low
    & Medium
    & High
    \\

    \parbox[c][\rh][c]{0.6cm}{\rotatebox{90}{\shortstack{Scene 10\\Camera 1}}} &
    \cropleft[height=\rh]{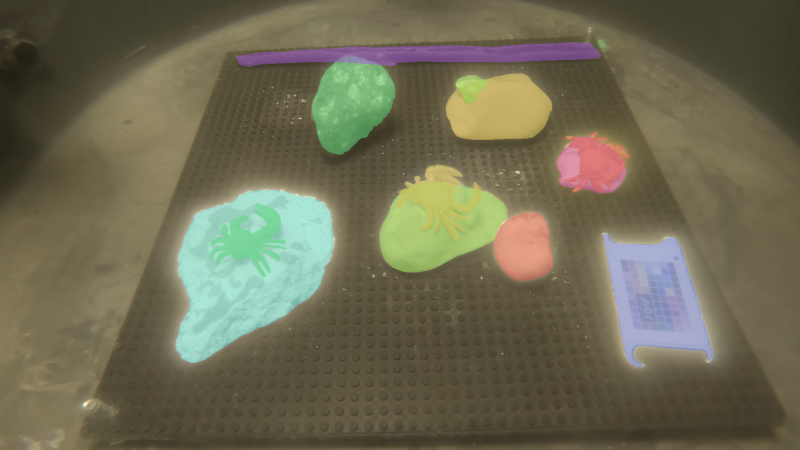} &
    \cropleft[height=\rh]{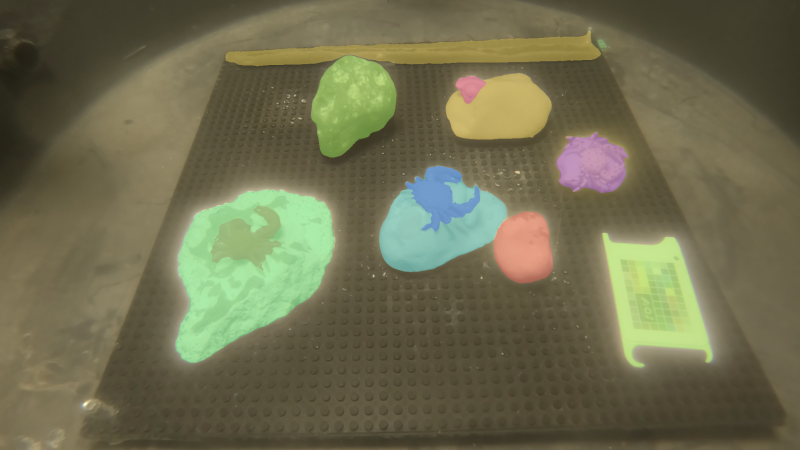} &
    \cropleft[height=\rh]{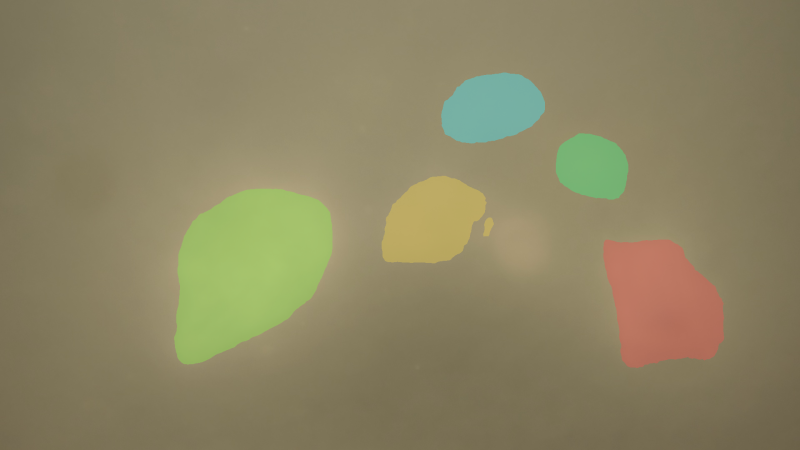} &
    \cropleft[height=\rh]{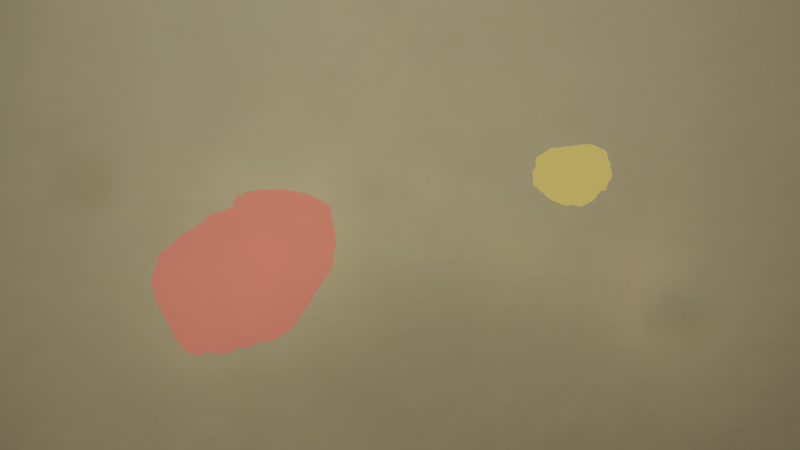}
    \\

    \parbox[c][\rh][c]{0.6cm}{\rotatebox{90}{\shortstack{Scene 15\\Camera 1}}} &
    \cropleft[height=\rh]{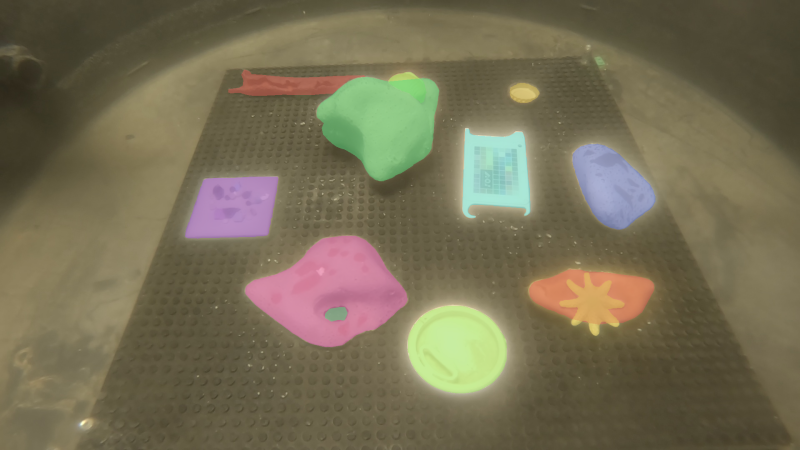} &
    \cropleft[height=\rh]{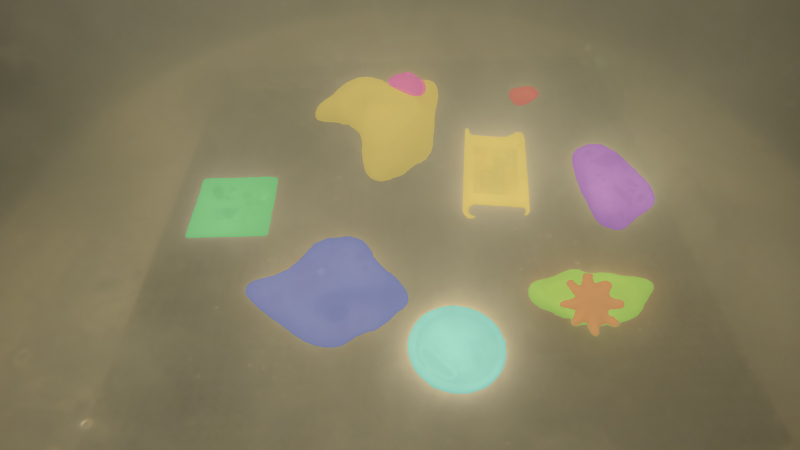} &
    \cropleft[height=\rh]{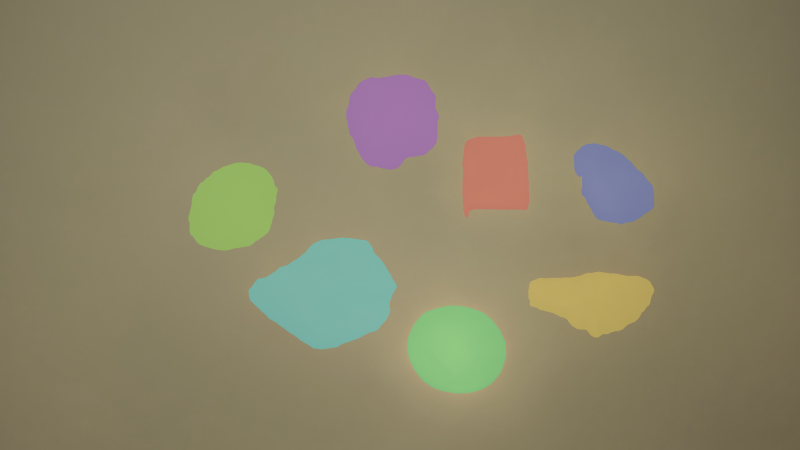} &
    \cropleft[height=\rh]{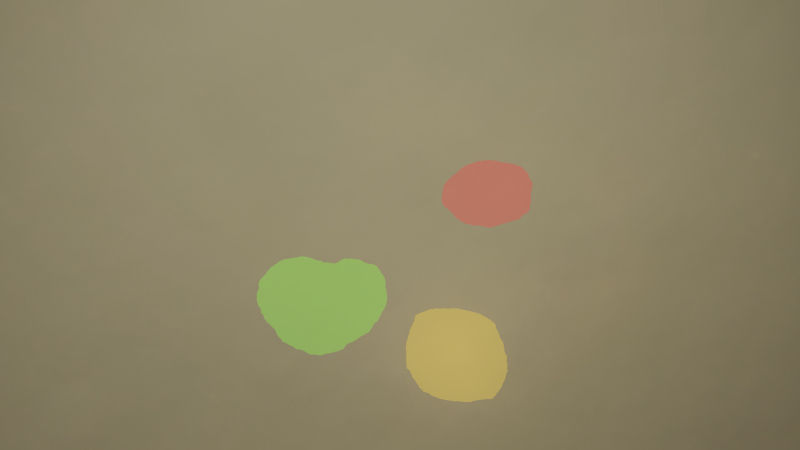}
    \\

    \parbox[c][\rh][c]{0.6cm}{\rotatebox{90}{\shortstack{Scene 20\\Camera 2}}} &
    \cropleft[height=\rh]{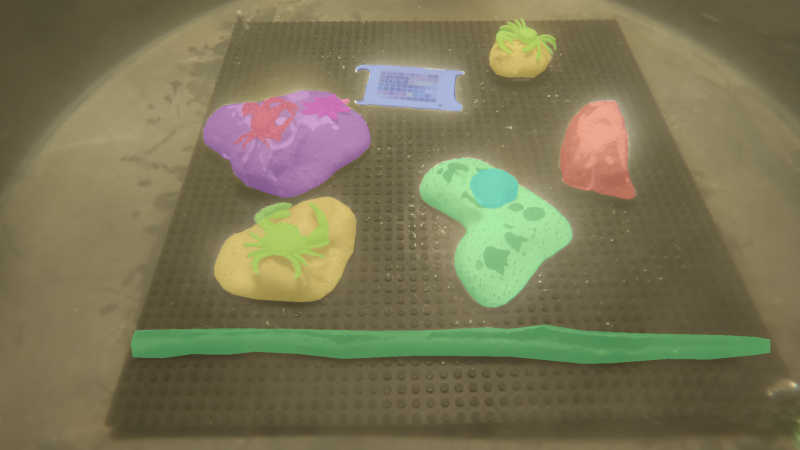} &
    \cropleft[height=\rh]{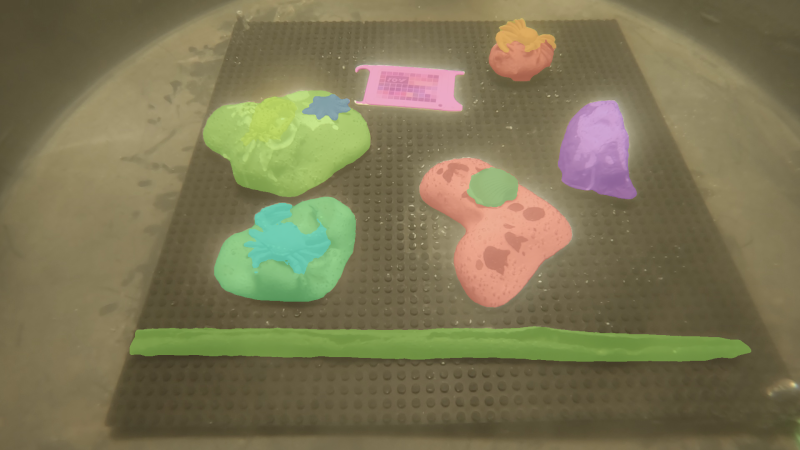} &
    \cropleft[height=\rh]{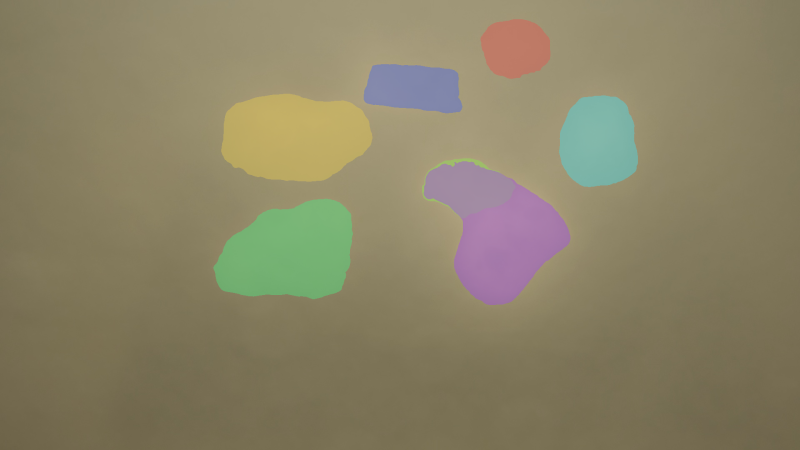} &
    \cropleft[height=\rh]{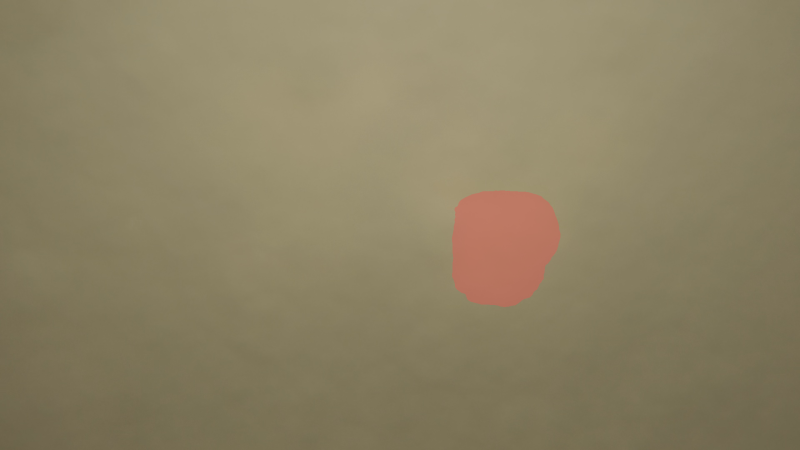}
    \\
};

\node[anchor=south, yshift=1mm] at ($ (m-1-3.north)!0.5!(m-1-5.north) $)
    {Turbidity-grading};

\draw[line width=0.3pt, -{Latex[length=2mm, width=1.2mm]}]
    ($ (m-1-3.north) + (0,1mm) $) --
    ($ (m-1-5.north) + (0,1mm) $);

\end{tikzpicture}

%% file: figures/predictions-examples/synthetic-prediction_examples.tex
\newcommand{\rh}{1.8cm} 

\newcommand{\img}[2][]{\cropleft[#1]{#2}}
\newcommand{\base}{figures/predictions-examples/mask2former}

\newcommand{\FNA}{Cam1_Exp10_Stage01_001}
\newcommand{\FNB}{Cam1_Exp15_Stage01_001}
\newcommand{\FNC}{Cam2_Exp20_Stage01_001}
\newcommand{\FND}{Cam4_Exp10_Stage01_001}

\newcommand{\FNAsfx}{4.704}
\newcommand{\FNBsfx}{7.778}
\newcommand{\FNCsfx}{1.208}
\newcommand{\FNDsfx}{4.859}

\begin{tikzpicture}[
    every node/.style={anchor=center, inner sep=0, outer sep=0},
    row sep=0pt,
    column sep=3pt
]

\matrix (m) [
    matrix of nodes,
    nodes={anchor=center},
    row sep=3pt,
    column sep=3pt
]{

      {}
    & \shortstack{Scene 10\\Camera 1}
    & \shortstack{Scene 15\\Camera 1}
    & \shortstack{Scene 20\\Camera 2}
    & \shortstack{Scene 10\\Camera 4}
    \\

    \parbox[c][\rh][c]{0.3cm}{\centering \rotatebox{90}{\small Ground truth}} &
    \img[height=\rh]{\base/gt/\FNA.png} &
    \img[height=\rh]{\base/gt/\FNB.png} &
    \img[height=\rh]{\base/gt/\FNC.png} &
    \img[height=\rh]{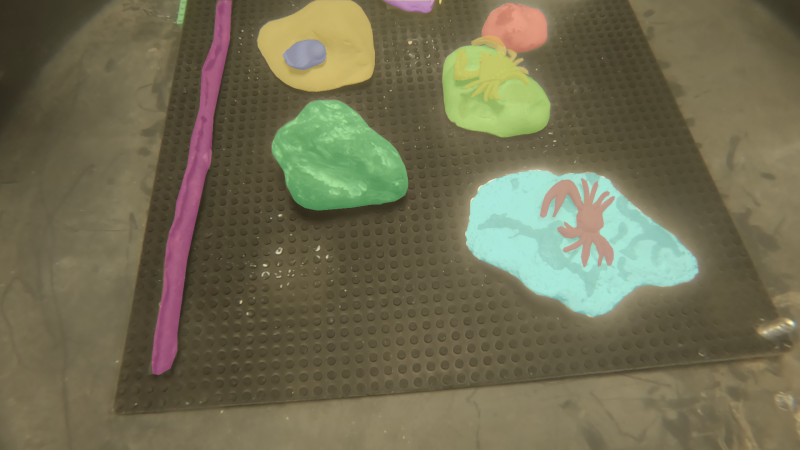}
    \\

    \parbox[c][\rh][c]{0.3cm}{\centering \rotatebox{90}{$Synth_1$}} &
    \img[height=\rh]{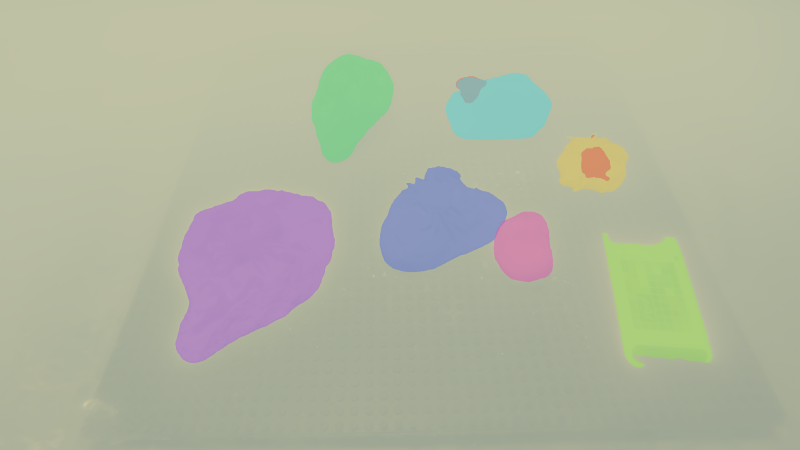} &
    \img[height=\rh]{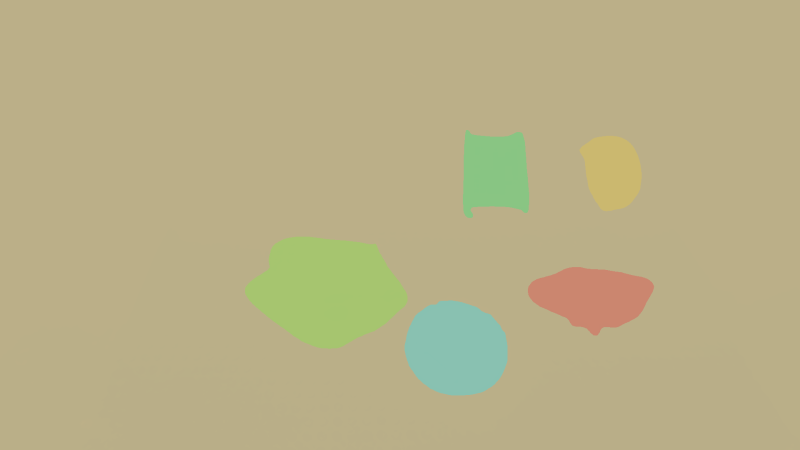} &
    \img[height=\rh]{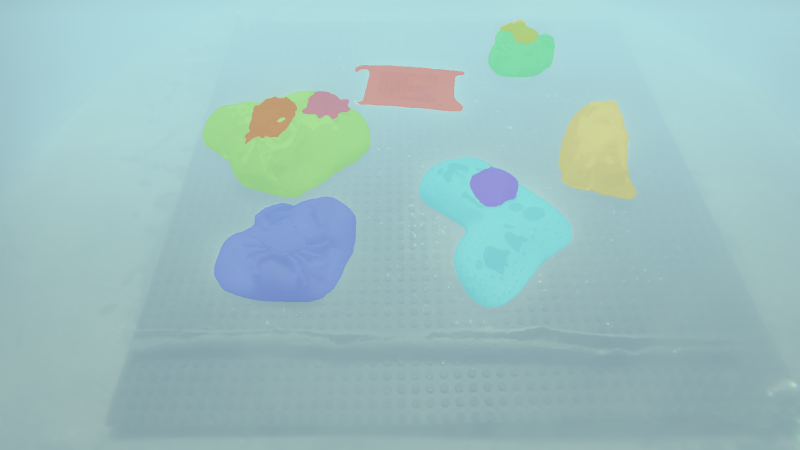} &
    \img[height=\rh]{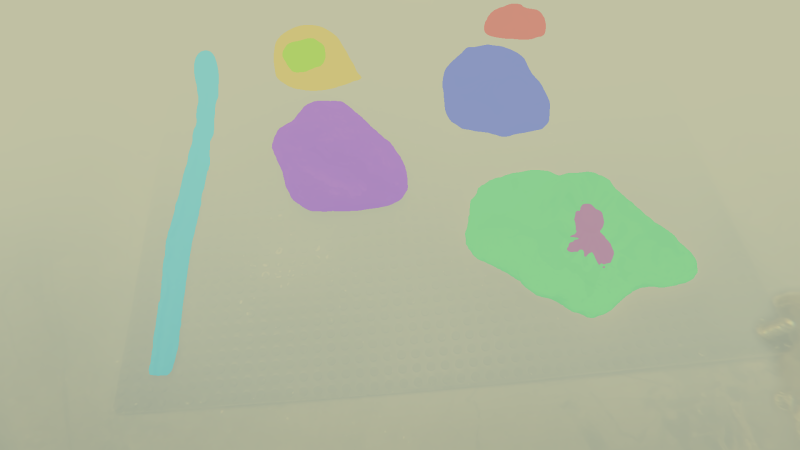}
    \\

    \parbox[c][\rh][c]{0.3cm}{\centering \rotatebox{90}{$Synth_2$}} &
    \img[height=\rh]{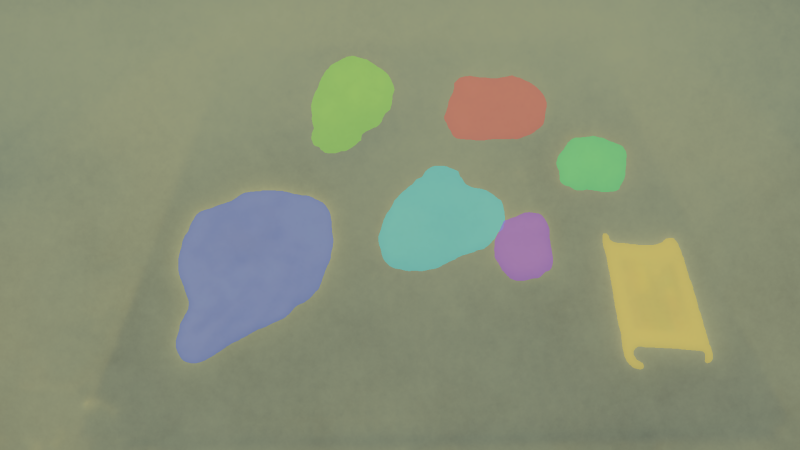} &
    \img[height=\rh]{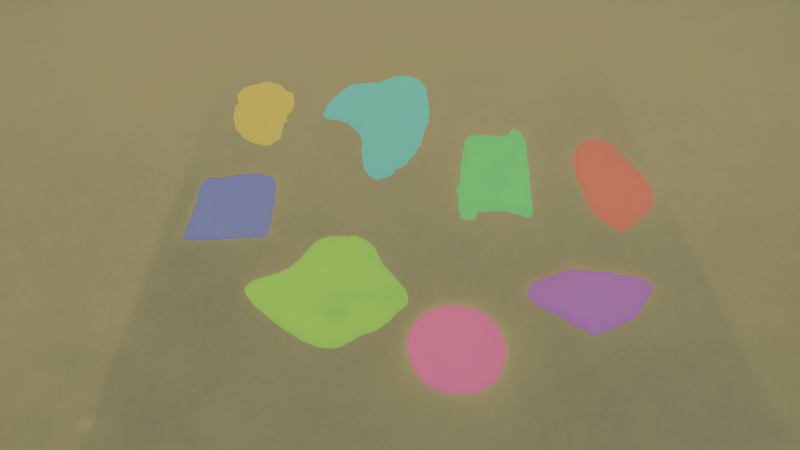} &
    \img[height=\rh]{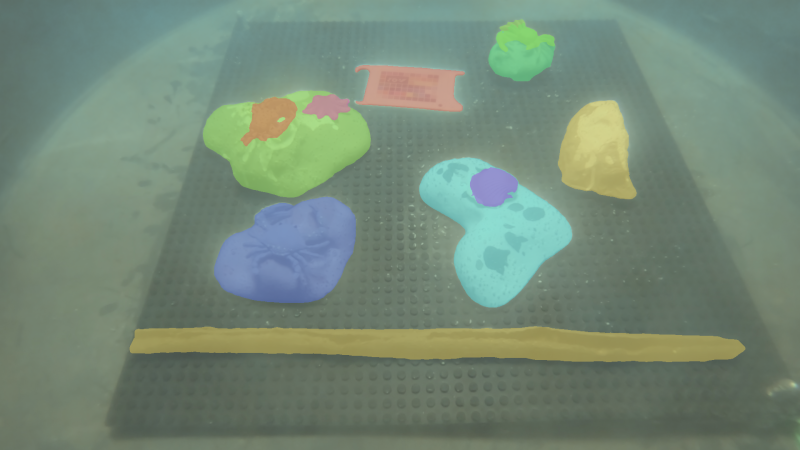} &
    \img[height=\rh]{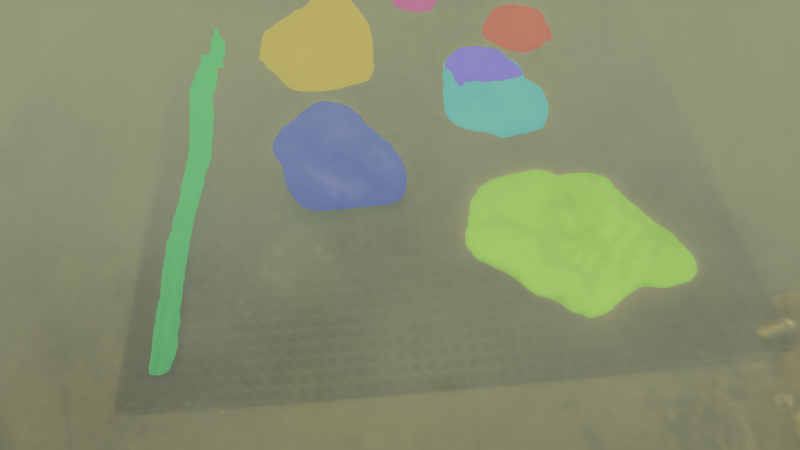}
    \\
};

\end{tikzpicture}